\documentclass[journal,compsoc]{IEEEtran}

\ifCLASSOPTIONcompsoc
  \usepackage[nocompress]{cite}
\else
  \usepackage{cite}
\fi
\ifCLASSINFOpdf
\else
\fi
\usepackage{graphicx}

\usepackage{xspace}

\usepackage{amsmath}
\usepackage{amssymb}
\usepackage{booktabs}
\usepackage{array}
\usepackage{floatrow}
\usepackage[table,x11names]{xcolor}
\usepackage{soul}
\usepackage{bm}
\usepackage[linesnumbered,ruled]{algorithm2e}
\usepackage{bbm}
\usepackage{url}
\usepackage[T1]{fontenc}
\usepackage[counterclockwise, figuresleft]{rotating}
\usepackage{multirow}
\usepackage{subfig}

\definecolor{LightCyan}{rgb}{0.88,1,1}
\DeclareMathOperator{\softmax}{softmax}

\DeclareMathOperator{\lstm}{LSTM}

\newcommand{\fnote}[1]{\textcolor[rgb]{0,0,0}{{#1}}}

\newcommand{\doublecheck}[1]{\textcolor[rgb]{0,0,0}{#1}}

\newcommand{\keypoint}[1]{\vspace{0.1cm}\noindent\textbf{#1}\quad}

\makeatletter
\DeclareRobustCommand\onedot{\futurelet\@let@token\@onedot}
\def\@onedot{\ifx\@let@token.\else.\null\fi\xspace}

\def\eg{\emph{e.g}\onedot} 
 \def\Ie{\emph{I.e}\onedot}
 
\def\etc{\emph{etc}\onedot} 
 
\def\etal{\emph{et al}\onedot}
\makeatother

\begin{document}
%
\title{Inverse Visual Question Answering: \\A New Benchmark and VQA Diagnosis Tool}

\author{Feng Liu,
Tao Xiang,
Timothy M. Hospedales, 
Wankou Yang, and
Changyin Sun

\IEEEcompsocitemizethanks{\IEEEcompsocthanksitem F. Liu, W. Yang, and C. Sun are with  
School of Automation, Southeast University, Nanjing, China, 210096. E-mail: \{liufeng, wkyang, cysun\}@seu.edu.cn
}
\IEEEcompsocitemizethanks{\IEEEcompsocthanksitem T. Xiang is with  
Queen Mary University of London, Mile End, London, E1 4NS.
E-mail: t.xiang@qmul.ac.uk
}
\IEEEcompsocitemizethanks{\IEEEcompsocthanksitem T. Hospedales is with the University of Edinburgh, 10 Crichton Street, Edinburgh, EH8 9AB.
E-mail: t.hospedales@ed.ac.uk
}
}
\IEEEtitleabstractindextext{%
\begin{abstract}
In recent years, visual question answering (VQA) has become topical. The premise of VQA's significance as a benchmark in AI, is that both the image and textual question need to be well understood and mutually grounded in order to infer the correct answer. However, current VQA models perhaps `understand' less than initially hoped, and instead master the easier task of exploiting cues given away in the question and biases in the answer distribution \cite{VQA2.0}. 
In this paper we propose the inverse problem of VQA (iVQA). The iVQA task is to generate a question that corresponds to a given image and answer pair. We propose a variational iVQA model that can generate diverse,  grammatically correct and content correlated questions that match the given answer. Based on this model, we show that iVQA is  an interesting benchmark for visuo-linguistic understanding, and a more challenging alternative to VQA 
because an iVQA model needs to understand the image better to be successful. 
As a second contribution, we show how to use iVQA in a novel reinforcement learning framework  to diagnose any existing VQA model by way of exposing its belief set: the set of question-answer pairs that the VQA model would predict true for a given image. This provides a completely new window into what VQA models `believe' about images. We show that existing VQA models have more erroneous beliefs than previously thought, revealing their intrinsic weaknesses.  Suggestions are then made on how to address these weaknesses going forward. 
 \end{abstract}

\begin{IEEEkeywords}
Inverse Visual Question Answering, VQA Visualisation, Visuo-Linguistic Understanding, Reinforcement Learning
\end{IEEEkeywords}}

\maketitle

\IEEEdisplaynontitleabstractindextext

%
\IEEEpeerreviewmaketitle

\section{Introduction}

\IEEEPARstart{A}{s} conventional object detection and recognition approach solved problems, we see a surge of interest in more challenging problems that should require greater `understanding' from computer vision systems. Image captioning \cite{showandtell}, visual question answering \cite{vqa_dataset}, natural language object retrieval \cite{KazemzadehOrdonezMattenBergEMNLP14} and `visual Turing tests' \cite{geman2015visualTuringTestVQA} provide multi-modal AI challenges that are expected to require rich visual and linguistic understanding, as well as knowledge representation and reasoning capabilities. As interest in these grand challenges has grown, so has scrutiny of the benchmarks and models that appear to solve them. Are we making progress towards these challenges, or are good results the latest incarnation of horses \cite{pfungst1991cleverHans,sturm2014horse} and Potemkin villages \cite{goodfellow2015adversarialExamples}, with neural networks finding unexpected correlates that provide shortcuts to give away the answer? 

Recent analyses of VQA models and benchmarks have found that the reported VQA success is largely due to exploiting dataset biases and cues given away in the question, with predictions being minimally dependent on understanding image content \cite{VQA2.0,vqahat,vqa_analysis1,jabri2016revisitVQA}. For example it turns out that existing VQA models do not `look' in the same places as humans do to answer the question \cite{vqahat}; they do not give different answers when the same question is asked of different images \cite{vqa_analysis1}; and they can  perform ``well'' given no image at all \cite{vqa_dataset,jabri2016revisitVQA}. Moreover, VQA model predictions do not depend on more than the first few words of the question \cite{vqa_analysis1}, and their success depends largely on being able to exploit label bias \cite{VQA2.0}. These observations have motivated renewed attempts to devise more rigorous VQA benchmarks \cite{VQA2.0}.

\begin{figure}[t]
\includegraphics[width=0.9\columnwidth]{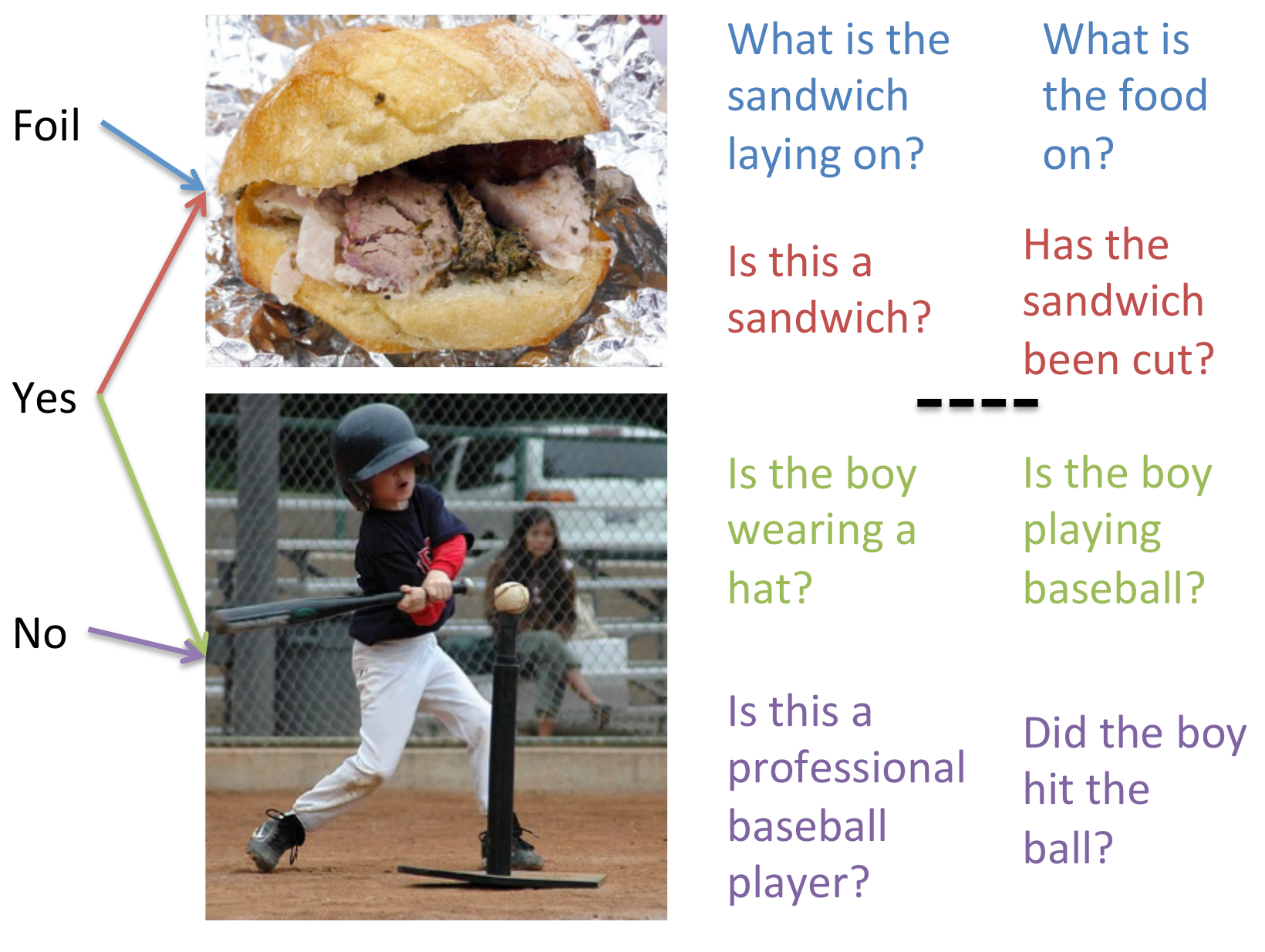}
\vspace{-0.2cm}
\caption{Illustration of iVQA: Input answers and images along with the top questions generated by our model. }\label{fig:overview}
\end{figure}

In this paper we explore the concept of inverse visual question answering (iVQA). The iVQA task is to input an image-answer pair, and then ask (output) a suitable question for which the given answer holds in the context of the given image. An illustration is shown in Fig.~\ref{fig:overview}. We explore two applications of the proposed iVQA task: Firstly, iVQA provides a new benchmark for testing multi-modal understanding which is more robust to `gaming' than conventional VQA. Secondly, we show how iVQA can be used as a tool to diagnose and evaluate existing VQA models.

\keypoint{A New Multimodal Intelligence Benchmark}  iVQA provides a distinct and interesting benchmark for multi-modal intelligence. This is because: (i) There may be less scope for an iVQA model to cheat through question bias than for VQA to score highly through answer bias (there is less question bias, and exploiting it is harder than for categorical answers) and (ii) The input answers themselves provide a very sparse cue in iVQA compared to questions in VQA. So there may be less opportunity to deduce the question from the answer alone in iVQA than there is to deduce the answer from the question alone in VQA. Thus succeeding at iVQA more clearly requires actual understanding of image content.

\begin{figure}[t]
\includegraphics[width=1.0\textwidth]{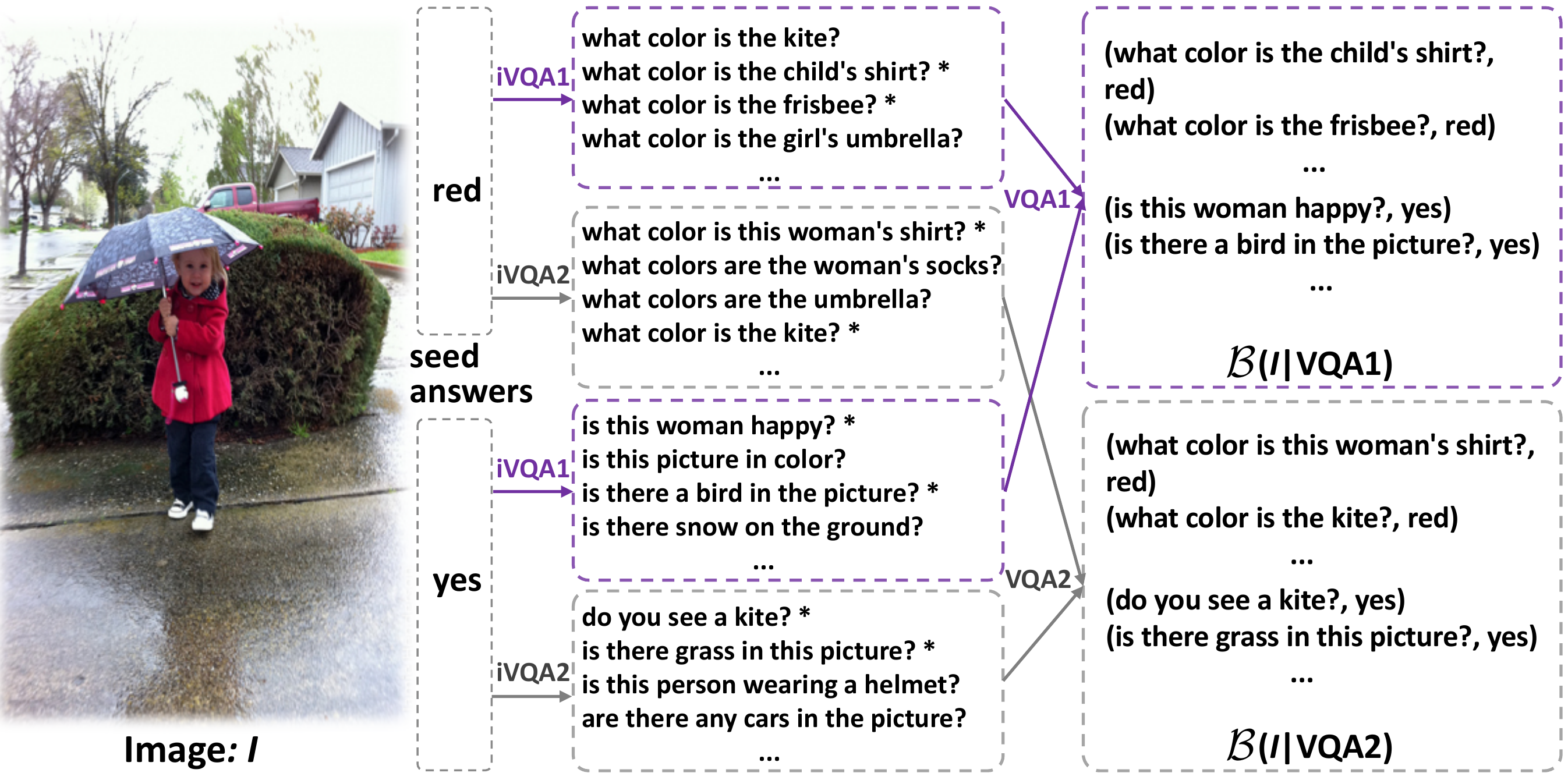}
\caption{Examples of belief sets of two  VQA models (VQA1 and VQA2) on image $I$. Candidates are  generated by corresponding RL-trained iVQA models (iVQA1 and iVQA2), and then verified by VQA models to build belief sets. 
}\label{fig:bs_frame}
\end{figure}

\keypoint{A VQA Diagnosis and Evaluation Tool} The standard approach to evaluating VQA has universally been to find out how well VQA models
predict the ground-truth answers of a set of pre-provided questions. This is a useful first step but, due to the closed-world assumption of this evaluation, we cannot validate the rationality of these models more comprehensively  without the restrictions imposed by the available question-answer annotations for a given image. In this work, we show how questions generated by our iVQA model can be used to expose the `belief set' of any given VQA model, which consists of {all the Q-A pairs that the VQA model would predict for a given image}, as illustrated in Fig.~\ref{fig:bs_frame}. This provides a completely new window `under the hood' of existing VQA models which allows us to understand how they are succeeding and failing to generalise. In particular, it allows us to \emph{diagnose} them by inspecting their belief sets for erroneous beliefs. This allows us to understand the specific weaknesses of different existing VQA models, and shows that existing VQA datasets are still inadequate (e.g., even models that score highly on the latest VQA datasets \cite{VQA2.0} are full of erroneous beliefs). 

By examining the beliefs of existing VQA models discovered by our approach, and comparing them to the annotations in the benchmark datasets, we find that newly discovered beliefs broadly fall into four categories: `complementary' -- true facts about an image that were not annotated during training, `rephrasing' -- questions that are differently phrased versions of those annotated during training, `irrelevant' -- image-irrelevant question and answer pairs \cite{quest_rel}, and `adversarial' -- image relevant questions for which the model believes the wrong answer. Counting the number of beliefs in each category provides a new way to evaluate VQA models' generalisation in a more open-ended way than existing closed-world benchmarks.

To satisfy both the above objectives, we propose a variational auto-encoder (VAE) \cite{journals/corr/KingmaW13} based iVQA model which includes randomness in question generation to produce diverse image+answer-conditional visual questions. For model diagnosis by belief-set computation, we further design a reinforcement learning (RL) based paradigm to generate visual questions that are tailored to a specific VQA model. The RL training is driven by a carefully designed reward scheme, where linguistic legality, uniqueness, and goal orientation are optimised to drive the policy evolution.

The contributions of this paper are  as follows: (1) The novel iVQA problem is introduced as an alternative benchmark of high-level multi-modal visuo-linguistic understanding. (2) A variational auto-encoder based iVQA model is proposed to generate diverse but image-answer-conditional visual questions. (3) We extend the iVQA model with RL training to intuitively diagnose and analyse VQA models via belief set generation. (4) 
We use the proposed belief-set generation approach to better evaluate VQA Models by observing the successes and failures of generalisation in visuo-linguistic understanding. (5) By combining belief sets of a number of state-of-the-arts VQA models, we contribute an extremely challenging VQA dataset upon which all existing VQA models tend to perform poorly, in order to complement existing VQA datasets. 

\vspace{-0.3cm}
\section{Related work}
\vspace{0.1cm}\noindent\textbf{Image Captioning}\quad Image captioning \cite{showandtell,feifeicap} aims to describe, rather than merely recognise objects in images. It encompasses a number of classic vision capabilities as prerequisites including object \cite{resnet} and action  \cite{actionRecognise} recognition, attribute description \cite{farhadi2009attrib_describe} and relationship inference \cite{lu2016visualRelationshipLanguage}. It further requires natural language generation capabilities to synthesise open-ended linguistic descriptions. Popular benchmarks and competitions have inspired intensive research in this area. Captioning models have explicitly addressed these sub-tasks to varying degrees \cite{feifeicap}, but the most common and successful approaches use neural encoders (of  images), and decoders (of  captions), with little explicit knowledge representation and reasoning \cite{showandtell,liu2017semanticRecurrent}. The iVQA task investigated here is related to captioning in that we aim to produce natural language outputs, but distinct in that the outputs are sharply conditioned on the required answer, as illustrated in Fig.~\ref{fig:overview}.

\vspace{0.1cm}\noindent\textbf{VQA Challenge}\quad Like captioning, VQA has gained attention as a synthesis challenge in AI, requiring both computer vision and natural language understanding to succeed \cite{vqa_dataset}. Given an image, and a natural language question about the image, a VQA system  produces an answer. Unlike other vision tasks (recognition, detection, description), the question to be answered in VQA is dynamically specified at runtime. Besides visuo-linguistic grounding, many VQA examples seem to require extra information not contained in the question or image, \eg,  background common sense about the world. Thus VQA is hoped to provide a long term goal for  AI-complete multi-modal intelligence.  However increasing scrutiny has shown that learning systems excel at finding shortcuts in terms of gaming the biases in answer distributions, and giveaway correlations \cite{vqa_analysis1,jabri2016revisitVQA}, leading to doubts about the level of visuo-linguistic intelligence implied by current results \cite{vqahat}. Although some benchmarks in principle require open-sentence answers, most answers are simple one-word outputs, and therefore the most common approach has been to formalise answer generation as a multi-class classification problem over the most frequent answers \cite{jabri2016revisitVQA,mcb}. Although successful, this is somewhat unsatisfactory as it is no longer an open-world challenge. In this paper we explore iVQA as a novel alternative open-world benchmark for visuo-linguistic understanding. 

\vspace{0.1cm}\noindent\textbf{VQA Models}\quad Existing VQA models are typically based on two-branch neural networks: a CNN image encoder and LSTM question encoder, which are merged before feeding to an answer decoder \cite{vqa_dataset,malinowski2015askYourNeurons}. Recently they have been enhanced through various means including better visuo-linguistic merging \cite{mlb,mcb}, varying degrees of explicit representation \cite{andreas2016neuralModuleNet}, reasoning with external knowledge bases \cite{wang2016fvqa}, and improving  visual encoding through attention \cite{mcb,xu2016vqaAttention}. With most recent models treating answer generation as a classification problem, these models cannot be directly modified for iVQA by simply swapping the answer and question encoder/decoder. {Moreover, VQA is closed-world in the sense that there is usually exactly one right answer for a given question, while iVQA is open-world in the sense that there are often multiple questions that have the same answer in one image. Our VAE-based solution addresses this property of the problem.}

\vspace{0.1cm}\noindent\textbf{VQA Diagnoses}\quad
Agrawal \etal \cite{vqa_analysis1} show that VQA models could fail on novel concepts and jump to conclusions before `reading' the whole question. Additionally, Abhishek \etal \cite{vqahat} show that the visual attention of VQA models is only weakly correlated with the human attention. The most relevant analysis  to ours is \cite{vqa_analysis2}, which analyses the behaviours of VQA models by measuring the saliency of different parts of the inputs, \eg, words in the question, pixels in the image. However, our method still significantly differs from this in that we study behaviour of VQA models by generating questions that reveal their beliefs. It can be used in conjunction with all the previous described techniques.

\vspace{0.1cm}\noindent\textbf{VQA Question Relevance}\quad Another line of VQA analysis \cite{quest_rel,quest_rel1} studied the significance of question relevance. They observed that VQA models make predictions based on language bias when presented with irrelevant/unseen questions, resulting in answer predictions ignoring the visual facts. Visual questions often contain  \emph{premise} -- objects and relationships that the question implies are in the image. VQA models should leverage valid premises, and be robust to false-premise questions. These are related to our work in that irrelevant and false-premise questions are studied in both works. However, crucially our approach \emph{generates} the set of questions that a VQA model actually believes (including irrelevant/false-premise questions). In contrast, prior work manually constructed questions to train a relevance detection model or boost existing VQA models -- {but these manually generated QA pairs may or may not be in the belief sets of existing VQA models. }

\vspace{0.1cm}\noindent\textbf{Related Challenges}\quad 
Our iVQA task is related to the recently studied task of \textit{visual question generation} (VQG): to generate a natural question that is related to the content of an image \cite{vqg}. Introduced in \cite{vqg}, VQG is further studied in  \cite{arxivVQG} where  DenseCap \cite{densecap} is used to generate region specific descriptions before being translated into questions, while Jain \etal \cite{diverse_vqg} proposed a VAE-based VQG model to improve question diversity. The iVQA models we propose are related to the VQG model in \cite{diverse_vqg} that both are VAE variants; however, they differ significantly in that our model uses answer as condition. VQG  is a pre-specified task unlike VQA and iVQA which are dynamically determined at runtime. Importantly, VQG is  easier in terms of required understanding. Since VQG is not required to be answer-conditional, it often generates very general open questions that even humans cannot answer. It does not need to understand the image clearly enough -- and ground the two domains richly enough -- to correctly condition the generated question on the answer.  Another relevant challenge is \textit{visually grounded conversation} (VGC) which  aims to generate natural-sounding conversations \cite{das2016visual,vqg_extend}. VGC typically starts with VQG, but the following responses and further questions are generated primarily following conversational patterns mined from social media textual data, and only loosely grounded on the image content as context. In contrast, the grounding on image content in much tighter in iVQA.

\vspace{0.1cm}\noindent \textbf{Neural Network Visualisation}\quad Visualisation of neural networks has  advanced greatly in recent years. Early approaches just found patches in dataset that maximise the activation of the corresponding neuron \cite{rcnn}. Zeiler \etal \cite{deconv} apply an additional deconvoluational operation on the patches for clear visualisation. Another type of approach is gradient-based \cite{vis1}, which visualises a neuron's preference starting from a randomly initialised patch. It iteratively updates the patch via gradient ascent under some regularisation. This idea is related to ours in that both maximise the score of a neuron by sample generation, but we learn a policy to generate sentences rather than generating image patches.

\vspace{0.1cm}\noindent \textbf{Adversarial Sample Analysis}\quad Besides visualising deep networks to understand them, another direction is to attack deep models by discovering adversarial samples. Szegedy \etal \cite{adv_attack} found that machine learning models are vulnerable to adversarial samples, where slightly perturbed inputs could be misclassified. This was further extended in \cite{adv_attack1}, which proposed a gradient based method for efficient adversarial sample generation. Adversarial samples have also been found for recurrent neural networks. Papernot \etal \cite{seq_attack} adapt this method to sequential inputs, where some words in a sequence are replaced according to the gradients. Our work differs from these in that entirely new sentences are generated to maximise the VQA model score instead of just perturbing the original sentences.

A preliminary version of this work was published in \cite{ivqa}. Compared with the earlier study, there are several key differences: 
\fnote{(i) In this study, we focus on the one-to-many mapping problem of iVQA, and proposed a variational iVQA model. It adopts a variational auto-encoder based framework, and introduces latent vectors to handle the uncertainty in questions, while the model in \cite{ivqa} mainly focuses on visual attention and ignores the uncertainty problem. 
(ii) We propose in this work a new concept of belief set, which is a set of question-answer pairs a VQA model believes to hold for a particular image, enabling us to actually see what a model believes. It is empowered by the reinforcement learning based training paradigm introduced in this study. The VQA model diagnosis ability of iVQA has not been considered in \cite{ivqa}.
(iii) Based on belief set extraction, we are able to provide novel means of evaluating existing VQA models and datasets, and draw useful insights from the evaluation results. In contrast, iVQA was not considered as an evaluation tool for VQA in \cite{ivqa}.}

\section{iVQA as a visuo-linguistic understanding benchmark}\label{sec:iVQAMethod}

\subsection{Problem formulation}
The problem of inverse visual question answering (iVQA) is to infer a question $q$ for which a given answer $a$ holds, in the context of a particular image $I$. Formally:
\begin{equation}
q^{*} = \max_{q} p(q|I, a;\theta_1),
\end{equation}
where $q$ is a sentence with words $(w_1,w_2,...,w_T)$, where $T$ is the number of words in $q$, and $\theta_1$ is the parameters of a question generation model. This is a challenging problem because the question generation process is a one-to-many mapping problem, and multiple questions can be valid for a particular image-answer pair. As shown in Fig.~\ref{fig:overview}, given the answer `Black', questions querying about the colours of `boy's shirt', `helmet', and `baseball bat' are all correct.

To explicitly control the diversity in question generation, we introduce an additional (noise) factor \fnote{$\bm{z}$} into the model. By varying \fnote{$\bm{z}$} we expect to perturb the distribution and attend to different objects or question topics, so the problem can be reformulated as follow:
\begin{equation}
q^{*} = \max_{q} p(q|\bm{z},I,a;\theta_1).
\end{equation}
The above formulation could be effectively modelled in a variational auto-encoder (VAE) based framework as follows.

\begin{figure}[t]
\includegraphics[width=0.9\textwidth]{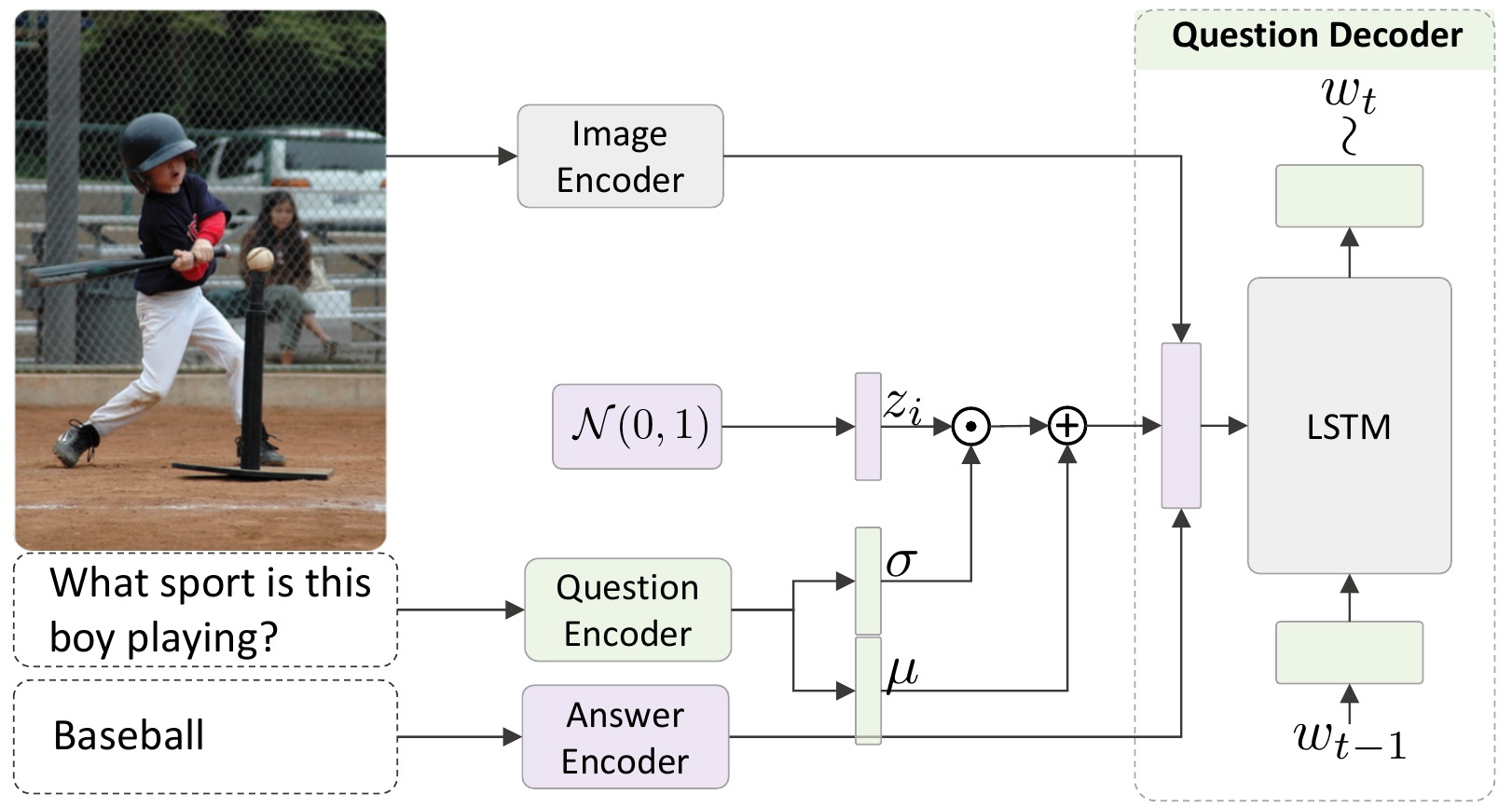}
\caption{Schematic of the proposed question generation model. The question encoder encodes the image and question to the mean and variance of a Gaussian distribution. Then the decoder \fnote{takes an image feature vector $\bm{f}_I$, an answer encoding $\bm{a}$, and a noise vector  $\bm{z}$ as inputs, and generates visual questions. The noise vector $\bm{z}$ is sampled from $\mathcal{N}(\bm{\mu},\bm{\sigma}^2\cdot \bm{1})$ during training, and $\mathcal{N}(\bm{0},\bm{1})$ for sampling}. }\label{fig:frame1}
\end{figure}

\subsection{Variational auto-encoders for iVQA} \label{sec:ivqa} 
The overall framework of an variational auto-encoder for question generation is shown in Fig. \ref{fig:frame1}. The model is  composed of an encoder $q_{\theta_1}(\fnote{\bm{z}}|I,q)$ and a decoder network $p_{\theta_2}(q|\fnote{\bm{z}},I,a)$, where $\theta_1$, $\theta_2$ are parameters of the encoder and decoder respectively. The encoder net encodes the question to distribution of latent factors $\fnote{\bm{z}}$, \fnote{parametrised by the mean $\bm{\mu}$ and variance $\bm{\sigma}^2\cdot \bm{1}$ of a Gaussian distribution}\footnote{\fnote{We use $\bm{1}$ to denote identity matrix to avoid notation confusion.}}. The decoder reconstructs the question from latent factors $\bm{z}$ and the conditioning image and answer. Its outputs are the parameters to a distribution \fnote{$\pi$} which describes the question. 
A sequence to sequence model is used as the backbone of our variational auto-encoder as we are dealing with {sequential data, i.e., questions}.

\noindent\textbf{Encoder}\quad For the encoder, an LSTM \cite{lstm_cell} is used, whose initial state is the encoded image features, as:
\begin{equation}
\begin{split}
\bm{c}_0^{e} &= \bm{0} \\
\bm{h}_0^{e} & =\bm{W}_{I}^{e}\bm{f}_I,
\end{split}
\end{equation}
where $\bm{W}_I^{e}$ is the weight of image embedding\footnote{Note that we skip the bias term throughout for simplicity.}, and $\bm{f}_I$ is the image feature. Then the LSTM network iteratively feeds in {question} tokens $w_t$ and updates its hidden states until the final state is reached. The state update is formulated as:
\begin{equation}
\bm{h}_{t}^{e},\bm{c}_{t}^{e} = \lstm^{enc}(\bm{x}_t,\bm{h}_{t-1}^{e},\bm{c}_{t-1}^{e}),
\end{equation}
where $\bm{x}_t$ is the word embedding of token $w_t$. Eventually, the parameters that describe the latent factors are computed based on the final hidden state $\bm{h}_{T}^e$ via:
\begin{equation}
\begin{split}
\bm{\mu} & = \bm{W}_{\mu}\bm{h}_{T}^{e} \\
\bm{\sigma} & = \delta(\bm{W}_{\sigma} \bm{h}_{T}^{e}),
\end{split}
\end{equation}
where $\bm{W}_{\mu}$ and $\bm{W}_{\sigma}$ are the weights for parameter prediction, and $\delta(\cdot)$ is the softplus activation function to ensure  non-negativity of the standard deviation.

\noindent\textbf{Decoder}\quad The decoder  samples a random vector $\bm{z}$ from the distribution $p_{\theta_1}(\fnote{\bm{z}}|I,q)$, based on which the question will be reconstructed. An LSTM network with the same structure is used for the decoding. Besides random vector $\bm{z}$, image feature $\bm{f}_I$ and answer encoding $\bm{a}$ are also used as inputs to the decoder. The inputs of the three modalities are fused as the initial hidden state of the decoder network, and the fusion is computed as:
\begin{equation} \label{eq:fuse}
\bm{h}_0^{d} = \bm{W}_{z}^{d}\bm{z} + \bm{W}_{I}^{d}\bm{f}_{I} + \bm{W}_{a}^{d}\bm{a},
\end{equation}
where $\bm{W}_{z}^{d}, \bm{W}_{I}^{d}, \bm{W}_{a}^{d}$ are corresponding fusion weights, while the cell state $\bm{c}_0^d$ is set to $\bm{0}$. The decoder then generates each word sequentially by taking the word embedding of the previous word $\bm{x}_{t-1}$ as input. Specifically, it is computed as:
\begin{equation} \label{eq:dec}
\begin{split} 
\bm{h}_{t}^{d},\bm{c}_{t}^{d} & = \lstm^{dec}(\bm{x}_{t-1},\bm{h}_{t-1}^{d},\bm{c}_{t-1}^{d})\\
\bm{v} & = \softmax(\bm{W}_{e}^{d} \bm{h}_{t}^{d}),
\end{split}
\end{equation}
where $\bm{v}$ is a vector which corresponds to the probability of taking every word in the vocabulary as output. The probability $p(w_t|w_{k<t};\fnote{\bm{z}},I,a)$ can be obtained by taking the $w_t$-th entry of $\bm{v}$, which is the probability of predicting word $w_t$ given previously generated tokens $w_{k<t}$.

\vspace{0.1cm}
\keypoint{Learning}
The VAE is trained by maximising the evidence lower bound (ELBO), which in our case is:
\begin{equation}
\max p_{\theta_1}(q|\fnote{\bm{z}},I,a)-KL(q_{\theta_2}(\fnote{\bm{z}}|I,q)||p(\fnote{\bm{z}}|I,q)),
\end{equation}
where the first term is the likelihood of generating the question $q$, which can be factorised as $\prod_{t=1}^{T}p(w_t|w_{k<t};\fnote{\bm{z}},I,a)$. The second term is the KL divergence between the encoder predicted distribution and its prior distribution. The prior distribution is chosen as the standard multivariate normal distribution $\mathcal{N}(\bm{0},\bm{1})$.

\vspace{0.1cm}
\keypoint{Inference} For open-ended question generation, multiple random vectors are sampled from the distribution $\mathcal{N}(\bm{0},\bm{1})$, then questions are generated from $p(q|\fnote{\bm{z}}_i,I,a)$ via beam search. Where a single question is required, we additionally select the one with maximum probability as the final output.

\subsection{iVQA evaluation}
We evaluate iVQA using three metrics including standard language-generation metrics, a new ranking-based metric, and a human validation study.

\vspace{0.1cm}\noindent\textbf{Linguistic Metrics}\quad
Standard linguistic measures \cite{DBLP:journals/corr/ChenFLVGDZ15} including  CIDEr, BLEU, METEOR and ROGUE-L  can be used to evaluate the generated questions. Given an (image, question, answer) tuple, we use the ground-truth question as the reference and compare the generated question based on the given image and answer. The similarity between the machine generated questions and the reference questions can be measured by these metrics. Even though generating humanlike questions is relatively easy, doing so in a correct image+answer conditional way  to get a high score is challenging, since the model has to capture all the semantic concepts and high-order interactions. 

\vspace{0.1cm}\noindent\textbf{Ranking Metric}\quad The linguistic metric approach has the issue that a generated question could score low despite perfectly corresponding to the provided image and answer. This is due to the one-to-many open-ended nature of iVQA -- it could ask a question that is correct but not in the annotated database. Therefore we develop a new ranking  based evaluation metric for the iVQA task. For an image-answer pair $(I,a)$, and a candidate question $q$, the conditioning score $p(q|I,a;\Theta)$ is used for ranking. If one of the correct (ground truth) questions is ranked the highest then this image-answer pair is regard as correct at Rank-1. In this way, accuracy over a testing set can be computed as the percentage of the times that correct questions are ranked at the top (denoted Acc.@1). Similarly, we can measure cumulative ranking accuracy at other ranks, e.g.,  Acc.@3 measures the  percentage of times  correct questions are ranked in top 3.  
This means that a model is not as severely penalised if it ranks the ground-truth questions highly, but its generated top ranked question is a novel (but potentially still correct) question. 
This metric is related to the multiple-choice setting of VQA \cite{vqa_dataset}. However we can use it  to reveal insights about model strengths and weaknesses by making specific choices of candidate questions in the  pool.

\vspace{0.1cm}\noindent\textbf{Question Pool}\quad
For a particular image-answer pair, the candidate ranking questions are collected from the following subsets. \textbf{Correct questions (GT)}: given image-answer pair $(I,a)$, the correct (ground truth) questions are defined as all the questions with answer $a$ in image $I$\footnote{There can be multiple correct questions corresponding to a given answer, since multiple questions may have the same answer for one image.}. \textbf{Contrastive questions (CT)}: these are questions associated with visually similar images to $I$ (including $I$) but having different answers. The similarity of the images is measured using the image CNN feature. \textbf{Plausible questions (PS):} These test whether the model can tell the subtle difference between questions and maintain grammatical correctness. They are obtained by randomly replacing one of the key words (e.g., verbs, nouns, adjective, and adverb) in the ground truth question. \textbf{Popular questions (PP):} Popular questions are chosen to be the most popular questions with the same answer type as $a$ across the whole dataset. These diagnose the extent to which the model is relying on label-bias. \textbf{Answer-related (RN):} these are chosen to be the random questions having answer $a$ but from other images. These diagnose the extent to which the model is relying on visual features, which did not always happen in VQA \cite{VQA2.0}. We manually checked all generated distractor questions, and removed any which were also correct for their corresponding image answer pairs.

\vspace{0.1cm}\keypoint{Human study} iVQA is open-ended and one-to-many in that there can be many correct questions for one image and answer. Therefore, given an image and an answer, the correct questions may not be annotated exhaustively. The proposed ranking metric computed on the selected question pool  alleviates the issue of comparing open-ended questions generated by an iVQA model to a fixed ground-truth set. However, it does not measure directly how `correct' the generated questions are, when they differ from the human annotations originally provided.  To evaluate iVQA in a way that awards `credit' for correct questions that are not annotated originally, we perform a human evaluation study. Image-answer pairs are randomly selected from the test set, and annotators assess the generated questions, scoring them from 0 (complete nonsense) to 4 (perfect). The mean score is used as the metric.

\section{Experiments on iVQA Benchmarking}
\subsection{Datasets and settings}
\keypoint{Dataset:} We repurpose the VQA dataset \cite{vqa_dataset} to investigate the iVQA task. The VQA dataset uses images from  MS COCO \cite{DBLP:journals/corr/ChenFLVGDZ15}, including 82,783 training, 40,504 validation  and 81,434 test images. Three question-answer pairs are collected for each image. Since the test set answers are not available, we adopt the commonly used off-line data split  in \cite{feifeicap,lu2017whenToLook} for image captioning: 82,783 images are used for training, and 5,000 each for validation and testing. 

\vspace{0.1cm}\keypoint{Question Pool:} For the proposed ranking accuracy metric, given each image-answer pair, the question pool  contains 24 questions, of which 1-3 are GT, 3-5 are CT (so that the total of GT+CT is 6), 6 are PP,  6 are PS, and 6 are RN .

\subsection{Baseline models}
\keypoint{Answer~only~(A):} It uses an LSTM encoder to encode tokenised answers to a fixed 512-dimensional representation, then an LSTM decoder to generate questions. 

\keypoint{Image only (I, VQG):} The visual only model is similar  to the GRNN model in \cite{vqg}. However, we use a more powerful image feature: the same \verb|res5c| feature of ResNet-152 \cite{resnet} used in our iVQA model and all other compared models. This feature is fed into an LSTM decoder as the initial state. 

\keypoint{Image+Answer Type (I+AT):}  VQG models \cite{vqg} generate questions purely based on visual cues. To make VQG more competitive in our answer-conditional iVQA setting, we also provide one-hot encoding of the answer \emph{type}. This hint helps VQG  generate the right question type (e.g.,`is' vs `what'). 

\keypoint{NN:} We adapt the nearest neighbour (NN) image captioning method \cite{nn}  to our problem. As iVQA is conditioned on both image and answer, we averaged the distance computed from both modalities for NN computation.

\keypoint{SAT:} Show attend and tell (SAT) \cite{sat} is a strong attentional captioning method. To provide a strong competitor to our approach, we modify  SAT to take input from both modalities by setting  the initial state of the decoding LSTM as the joint embedding of image and answer.

\keypoint{VQG+VQA:} The VQG+VQA baseline uses the VQG model above to generate question proposals from the image, and then uses VQA to select the question with maximum conditioning score. We use VQG to generate 10 candidates for each image for VQA re-ranking. The retrained multi-modal low-rank bilinear attention network \cite{mlb} is used as the VQA model.

\keypoint{Ours:} {Our model encodes images with a global semantic feature vector, which is obtained following \cite{liu2017semanticRecurrent} by learning a concept predictor on the training split with the 1,000 most frequent caption words as concepts.}

\begin{figure*}[t]
\includegraphics[width=0.85\textwidth]{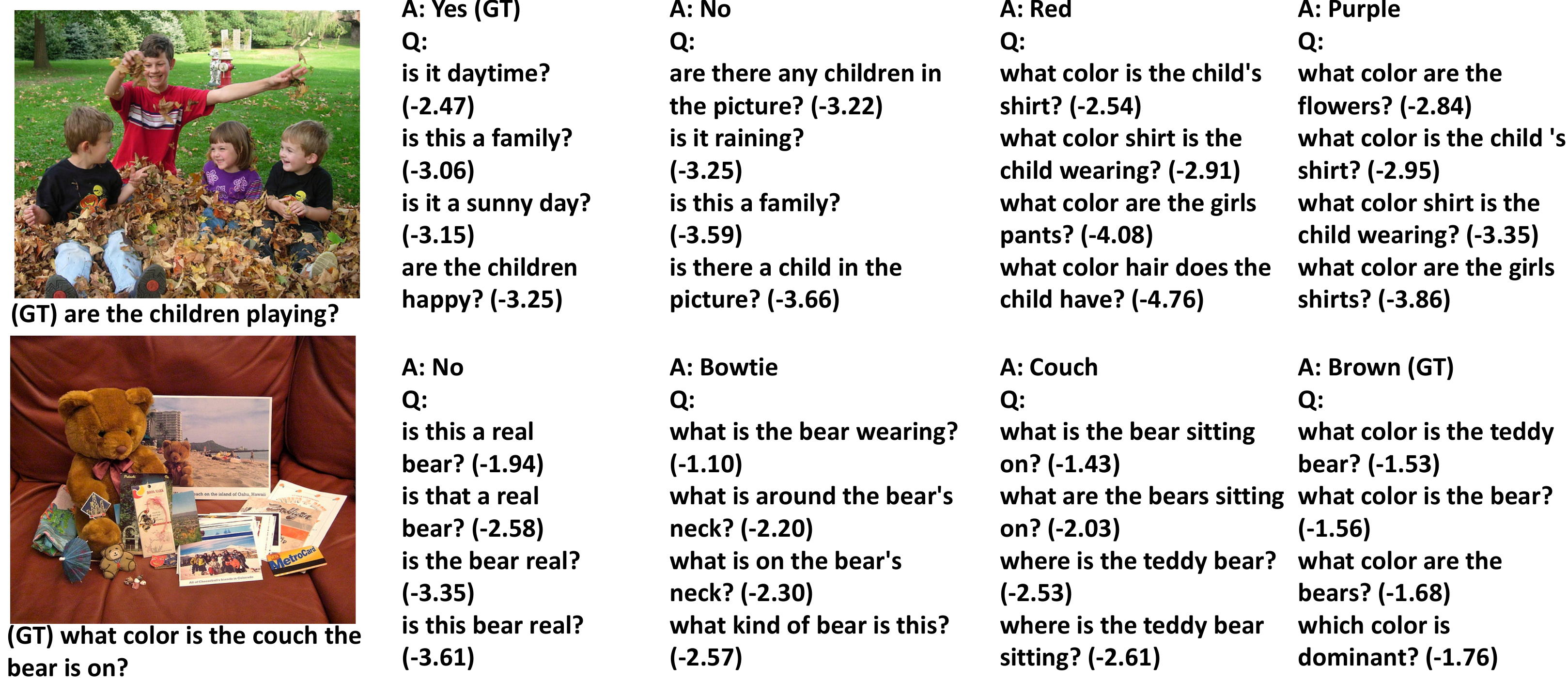}
\caption{Qualitative results of iVQA. Larger numbers in brackets mean higher confidence. }\label{fig:qualitative}
\end{figure*}

\begin{table*}[!thb]
\footnotesize
\begin{center}
\begin{tabular}{@{} l|c|cccc|c|c|cc|c @{}}
\toprule
 	& CIDEr	& BLEU-4	& BLEU-3	& BLEU-2	& BLEU-1	& ROUGE-L	& METEOR	& Acc@1	& Acc@3 & Human*\\ 
\midrule 
A	& 0.952	& 0.146	& 0.192	& 0.265	& 0.371	& 0.408	& 0.161	& 14.589	& 28.795 & 2.00 \\ 
I	& 0.652	& 0.086	& 0.121	& 0.179	& 0.280	& 0.310	& 0.117	& 13.012	& 28.644 & 2.10\\ 
I+AT	& 0.904	& 0.122	& 0.164	& 0.234	& 0.350	& 0.397	& 0.151	& 20.277	& 36.134 & 2.70\\ 
\midrule
NN	& 1.372	& 0.175	& 0.223	& 0.294	& 0.404	& 0.428	& 0.183	& 26.783	& 48.755 & 3.11\\
SAT	& 1.533	& 0.192	& 0.241	& 0.311	& 0.417	& 0.456	& 0.195	& 29.722	& 48.118 & 3.30\\  
VQG+VQA	& 1.110	& 0.147	& 0.193	& 0.261	& 0.371	& 0.396	& 0.165	& 16.529	& 41.655 & 2.85\\ 
\midrule
Ours	& {\bf 1.682}	& {\bf 0.205}	& {\bf 0.253}	& {\bf 0.320}	& {\bf 0.421}	& {\bf 0.466}	& {\bf 0.201}	& {\bf 30.814}	& {\bf 49.653} & {\bf 3.37}\\ 
\bottomrule
\end{tabular}
\end{center}
\vspace{-0.4cm}
\caption{Overall question generation performance on the testing set.}
\label{tab:main_table}
\end{table*}

\begin{figure*}[t]
\begin{center}
\begin{tabular}{@{}l@{}l@{}l@{}l@{}l@{}l@{}l}
\includegraphics[width=0.16\linewidth]{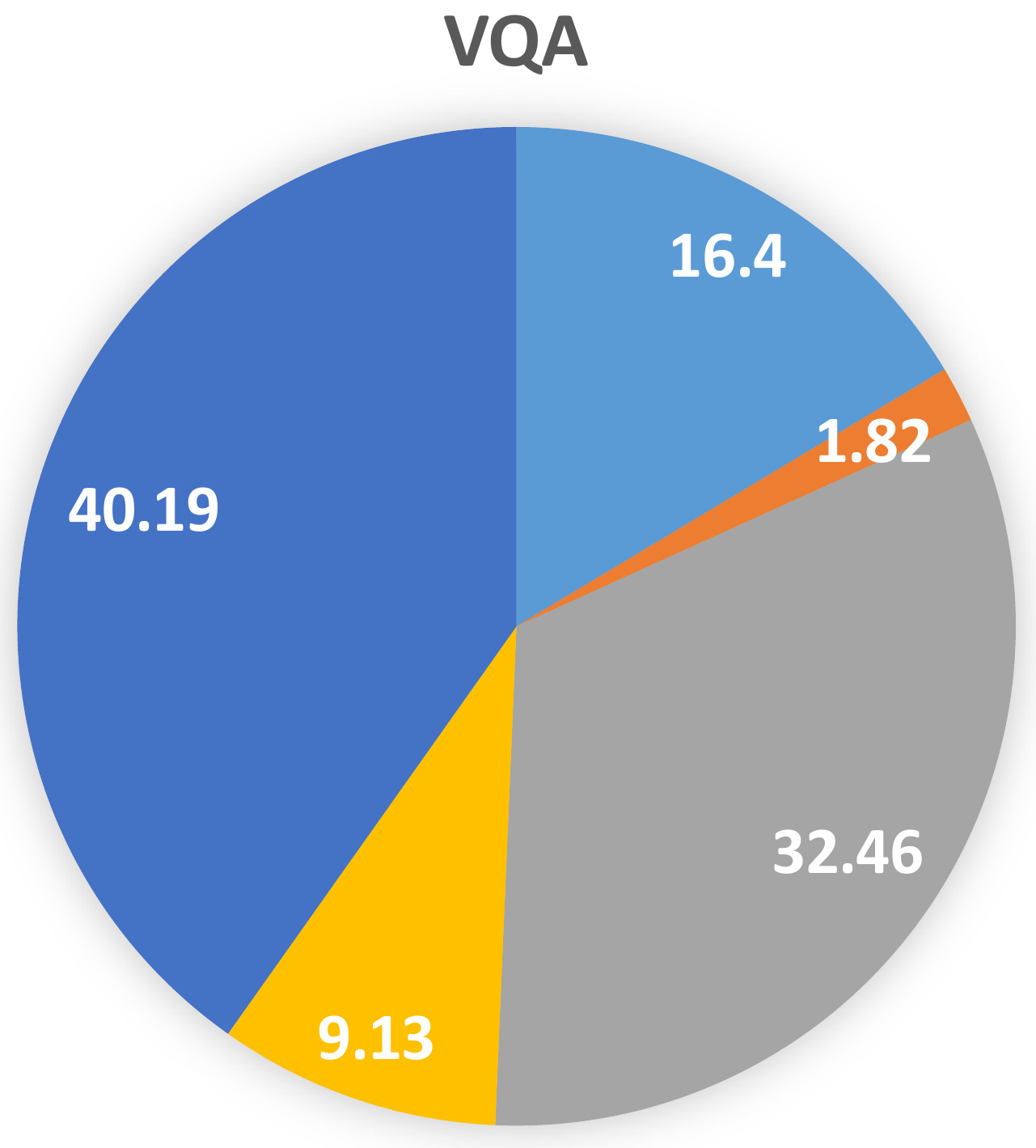}
& \includegraphics[width=0.16\linewidth]{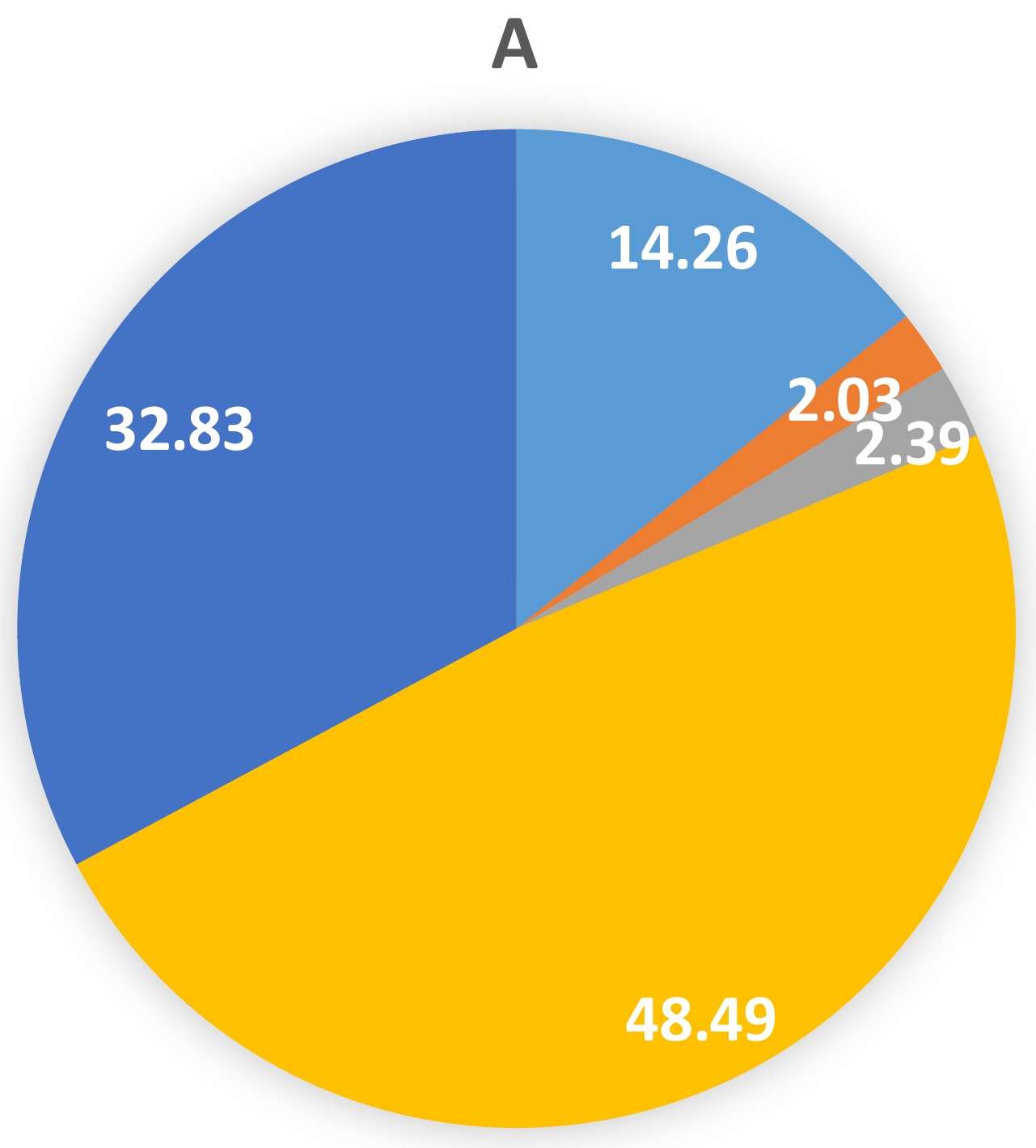}
& \includegraphics[width=0.16\linewidth]{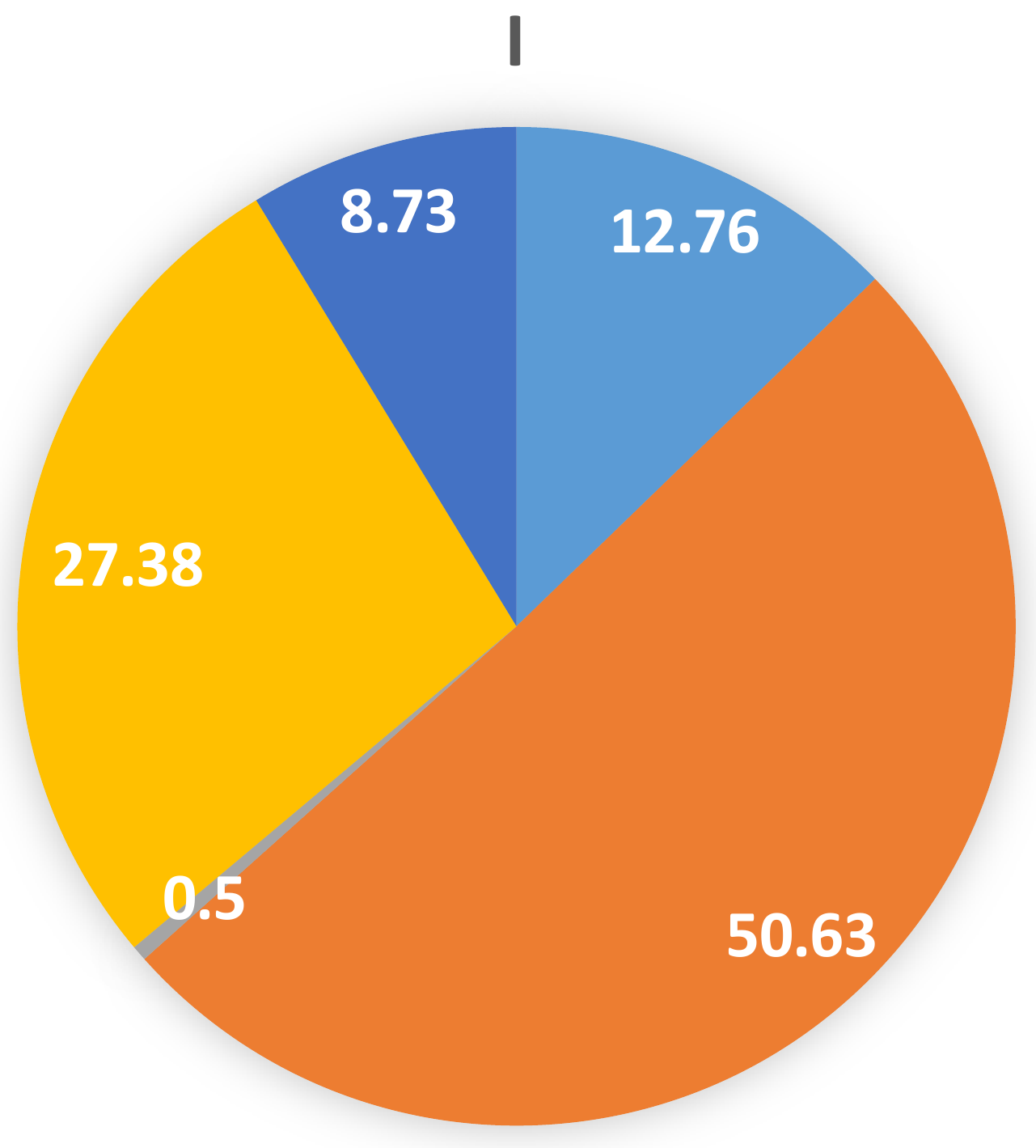}
& \includegraphics[width=0.16\linewidth]{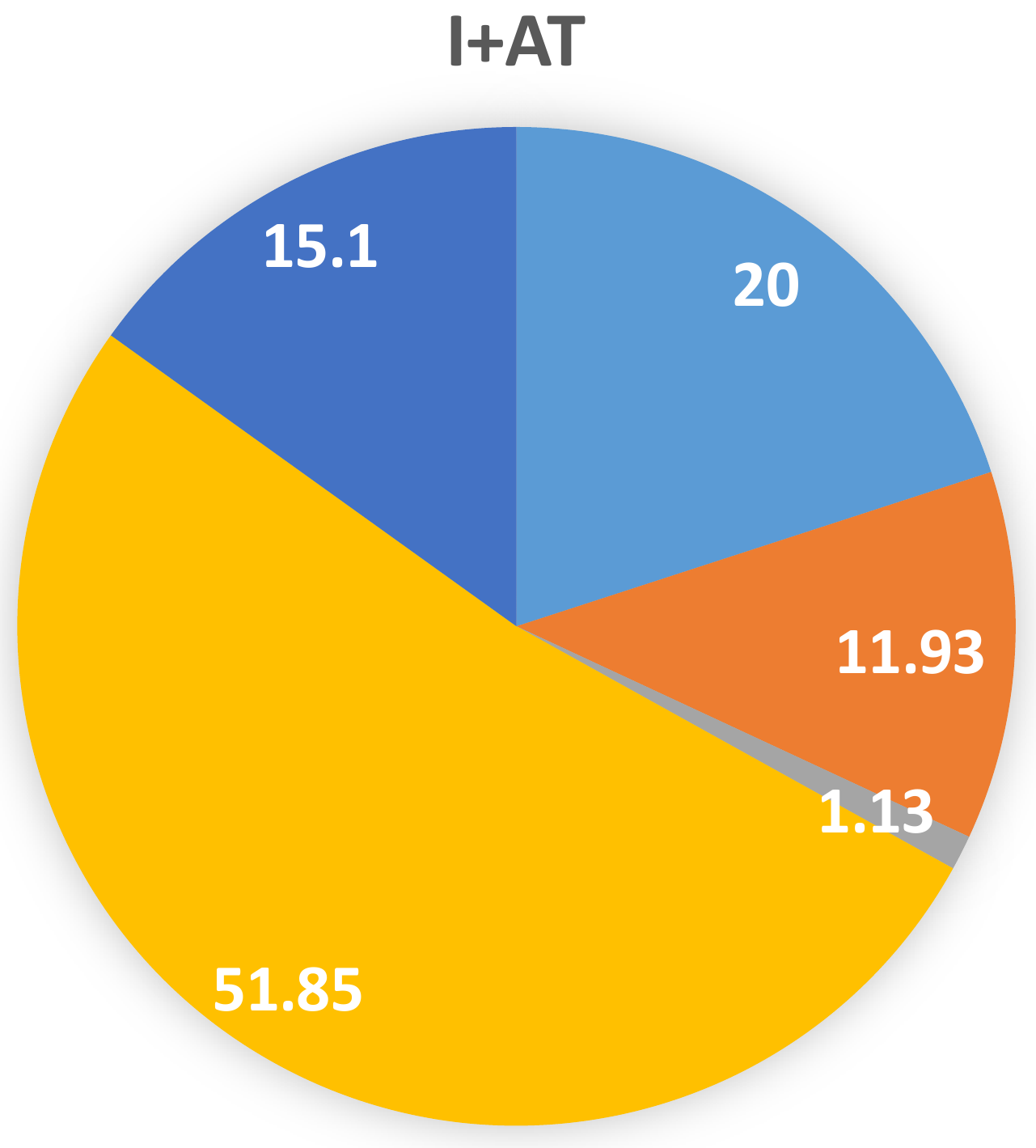}
& \includegraphics[width=0.16\linewidth]{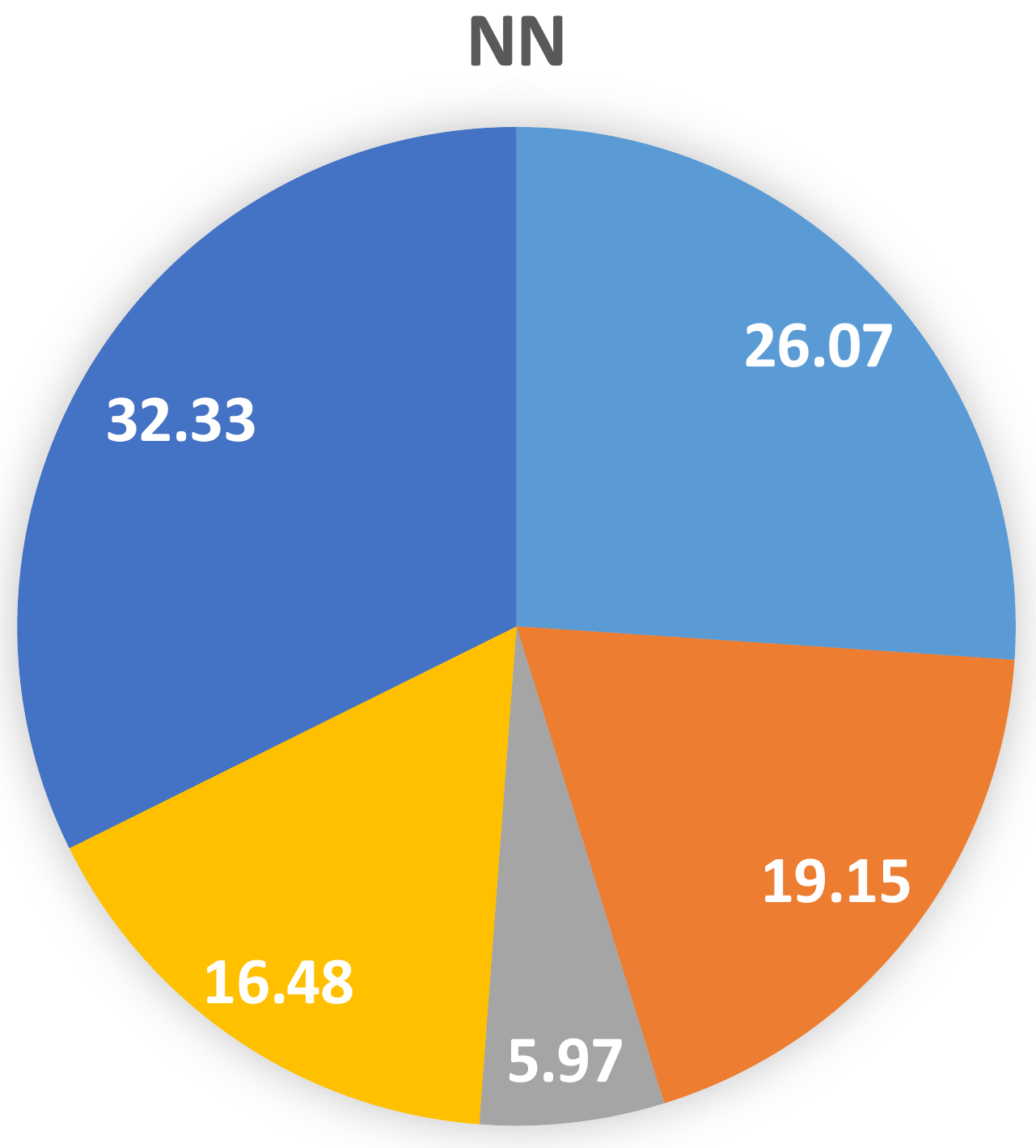}
& \includegraphics[width=0.162\linewidth]{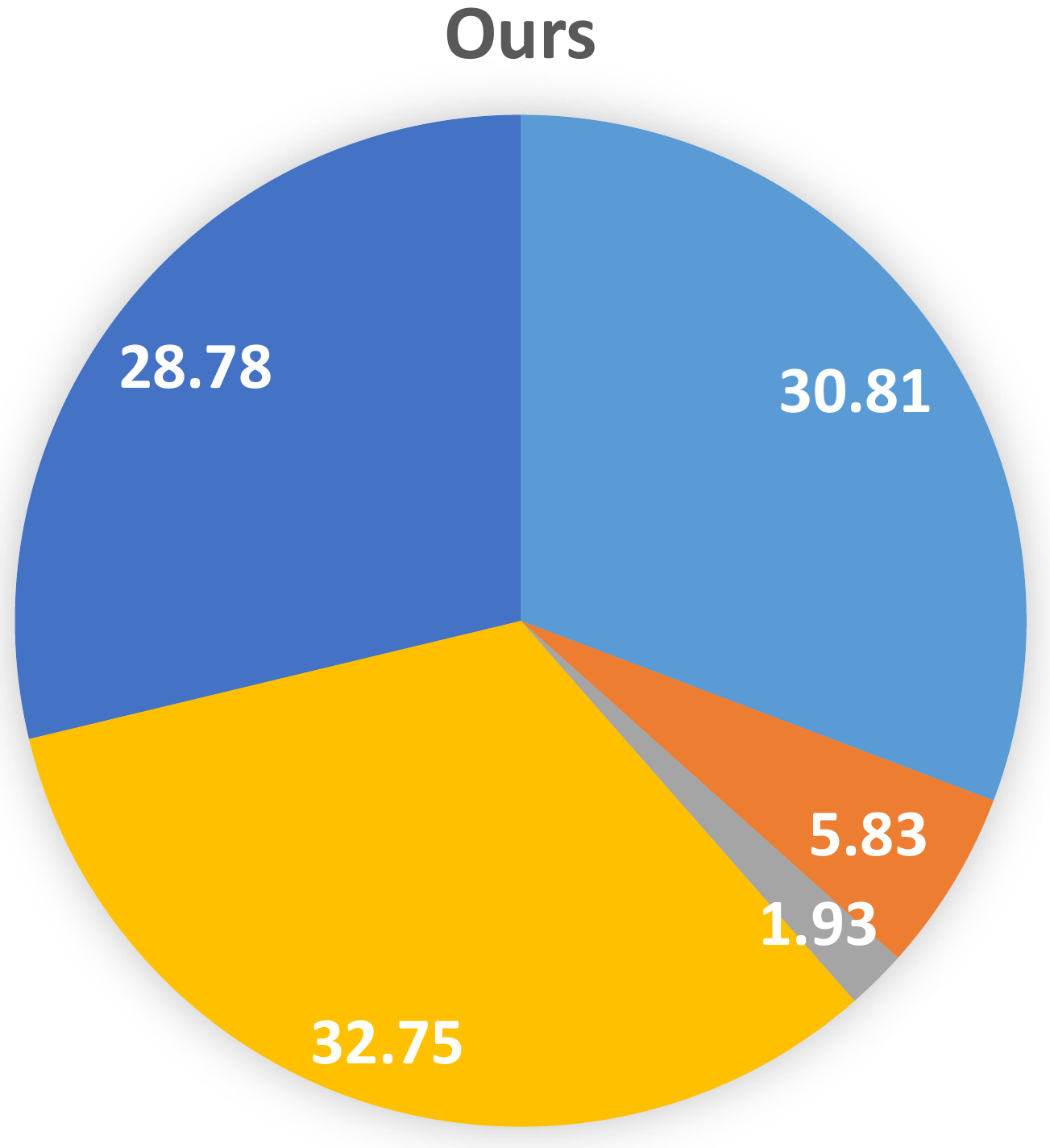}
& \ \includegraphics[width=0.05\linewidth]{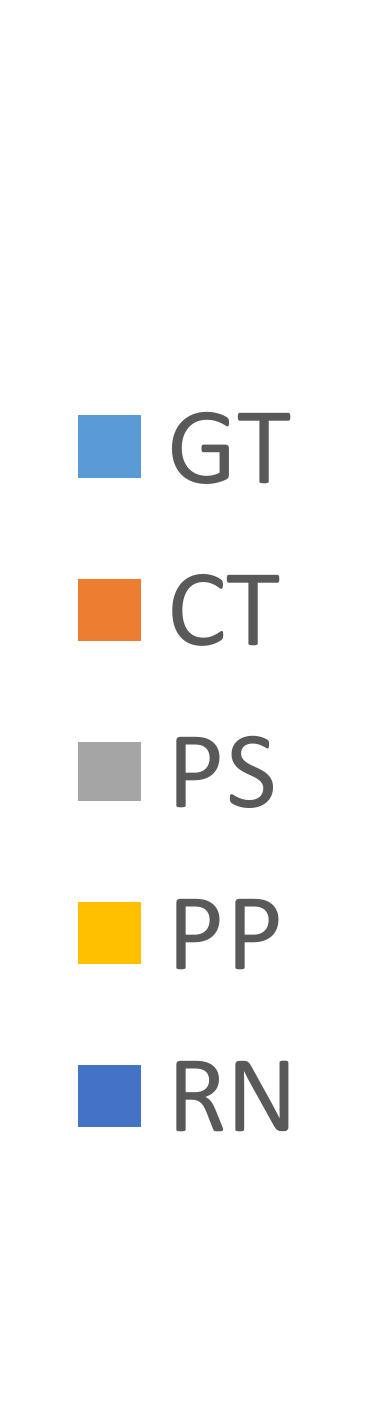}\\
\end{tabular}
\end{center}
\vspace{-0.4cm}
\caption{Comparison of effects of different distractors on different models on the test set.} \label{fig:distractorTypes}
\end{figure*}

\subsection{Results}
\keypoint{Overall} In the first experiment we report the overall iVQA performance on the test split. The results are shown in Table~\ref{tab:main_table} with both the standard linguistic metrics, as well as our ranking accuracy metric. From the table, we can make the following observations: (i) Unlike VQA the margin between the no-image case (A), and the full model (Ours) is dramatic. The ranking accuracies are more than doubled, and the language metrics show similarly striking improvements. This demonstrates that unlike conventional VQA \cite{VQA2.0}, the `V' does matter in iVQA.
(ii) The margin between the image + answer type (I+AT) and image-only (I) setting exists, but is not too significant. This shows that while it is a useful hint  for an iVQA model to know the question type, it still really needs the actual semantic answer to generate the right questions. E.g., rather than just knowing that it was counting something (answer type), the model does need to know \emph{how many objects were counted} (answer) in order to generate the right question specifying what object type needs to be counted -- as there may be other objects that could be counted.  
(iii) The margins between the (I) and (I+AT) cases and the full model are also striking. This demonstrates that as a test of multi-modal intelligence, iVQA reassuringly requires both modalities in order to do well. 
(iv) VQG+VQA indeed performs better than the vanilla VQG model by making the generated question more answer conditional, but it is still weaker than the captioning adapted models (NN and SAT) or  ours. The reason is that due to the language bias, it is difficult for the VQA model to identify the right question from multiple candidates where the same answer applies.
(v) The  captioning adapted models (NN and SAT) perform well, but are still inferior to the proposed model which is specifically designed for iVQA.

\vspace{0.1cm}\keypoint{Human Study} The human study is applied on a subset of 300 samples, and evaluates the models in a way that is fully robust to open-ended question generation. Results  in Table \ref{tab:main_table} show that (i) our model performs the best among all competitors; (ii) the human study scores are highly correlated with the proposed ranking metric. Specifically, the Pearson correlation coefficient between manually labelled scores and the proposed acc@1 and acc@3 metrics are 0.917 and 0.981 respectively, while the best performing linguistic measure (CIDEr)  only reaches 0.898. Thus our ranking metric is a reasonably accurate yet cost-effective alternative to the more expensive human evaluation.

\vspace{0.1cm}\keypoint{Qualitative Results} Examples of questions generated by  our model are shown in Fig.~\ref{fig:qualitative}. The results illustrate a few interesting points: (i) The generated questions are highly conditional on both images and answers. Particularly, the same answers generate different questions for different images, unlike in VQA \cite{vqa_analysis1}; and the same images generate very different questions when paired with different answers, showing  richer reasoning than in VQG \cite{vqg}. (ii) Unlike VQA, there are multiple reasonable questions that correspond to one image-answer pair. This is both due to alternative phrasing of the same question {(`\emph{where is the bear sitting?}','\emph{where is the teddy bear?}',`\emph{where is the teddy bear sitting?}')}, as well as multiple semantically distinct questions having the same answer {(e.g., `\emph{are the children happy?}',`\emph{is it daytime?}',`\emph{is it a sunny day?}')}. Since the annotation is not exhaustive,  standard linguistic metrics could be misleading: the generated questions can be correct but unannotated in the dataset. Our proposed ranking metric is more robust to this, as models are only scored according to how plausibly they rate the true question, rather than whether their open-world estimate of the question matches  annotated ground-truth.  Our human study evaluates the methods in a way that credits open-world question generation.  The open-ended question generation formulation of our variational iVQA model means it is straightforward to \emph{sample} the  distribution over questions given images and answers, in order to explore the model's beliefs -- a capability which we will exploit in the next section.

\keypoint {Analysis by Failure Type}
The proposed  new evaluation metric enables us  to understand the mistakes each model makes. The results in Fig.~\ref{fig:distractorTypes} show the Rank-1 predictions of each model (highest scoring question) broken down according to the category of that prediction. It again shows the superior performance of our iVQA model (ours): 30.81\% of the top ranked predictions are correct, which almost doubles that of the VQA model (16.4\%). But the main objective here is to analyse which distractor types are mistakenly ranked highly: 
(i) Without being able to condition on the answers, the image-only  method (I) makes predictions dominated by contrastive (CT) distractor questions taken from similar looking images. 
(ii) The answer-only (A) and image+answer type (I+AT) methods make predictions that are dominated by distractor questions of the popular (PP) type. This confirms that these models rely on the unconditional distribution of label statistics in order to solve iVQA. 
(iii) The VQA-based baseline instead is dominated by answer-related (RN) distractor. This suggests that the VQA model fails to correctly take into account image context (the `V' is not being accounted for \cite{vqa_analysis1,VQA2.0}), and is simply picking any question that generates the corresponding answer independently of the required contingency of the answer on the image context. 
(iv) The nearest neighbour (NN) approach performs well and has an evenly distributed set of error types, but it is weaker than the proposed in correctly capturing the visual and answer conditions. It is reflected on the larger portion of CT and RN errors. 
(v) Our full model \fnote{(ours)} has the largest fraction of correct predictions and manage to suppress the plausible (PS) and contrastive (CT) errors. However it still makes  mistakes ranking some popular (PP) questions at the top. This suggests that PP is a challenging distractor type and there is still some overfitting to the biased question distribution in the dataset \cite{VQA2.0}.

\begin{table}[h]
\footnotesize
\begin{center}
\begin{tabular}{@{} l|c|ccc @{}}
\toprule
 	& split	& Prior	& Language	& Language+Visual \\ 
\midrule 
iVQA (acc@1)	& test	& 3.94	& 14.59	& 28.44 \\ 
VQA (accuracy)	& test-dev	& 29.66	& 48.76	& 57.75 \\ 
\bottomrule
\end{tabular}
\end{center}
\vspace{-0.2cm}
\caption{VQA vs iVQA in terms of bias-based gameability.}\label{tab:vqa_ivqa}
\end{table}

\keypoint{Contrasting VQA and iVQA as benchmarks} \doublecheck{Finally, we discuss iVQA's interest as a benchmark compared to the conventional VQA. Two of the main kinds of bias that a VQA/iVQA model could use to cheat the benchmark are the output \emph{Prior bias}   (Ignore both inputs and predict only the most likely answer on VQA; use question frequency in iVQA), and \emph{Language bias} (ignore the image and use only the input -- question for VQA, answer for iVQA -- to predict the output). A good multimodal intelligence benchmark should require understanding and mutual grounding of both modalities, and should be hard to game by exploiting those biases. To analyses these issues we compare performance on iVQA and VQA benchmarks using Prior-alone and Language-alone (LSTM Q for VQA and Answer only for iVQA) baselines versus the full multi-modal model in each case (DeeperLSTM+Norm I for VQA, and I+A model for iVQA).
The results in Table \ref{tab:vqa_ivqa} show that for VQA the bias-based baselines approach the performance of a full multi-modal model much more closely than the corresponding baselines do for iVQA. This suggests that VQA is easier to `game' (achieve an apparently high score without any image understanding or multimodal grounding), compared to iVQA. Thus we propose that iVQA makes a distinct and interesting benchmark for multimodal intelligence.}

\section{iVQA as a VQA Diagnosis and Evaluation Tool}

\begin{figure*}[h]
\includegraphics[width=0.95\textwidth]{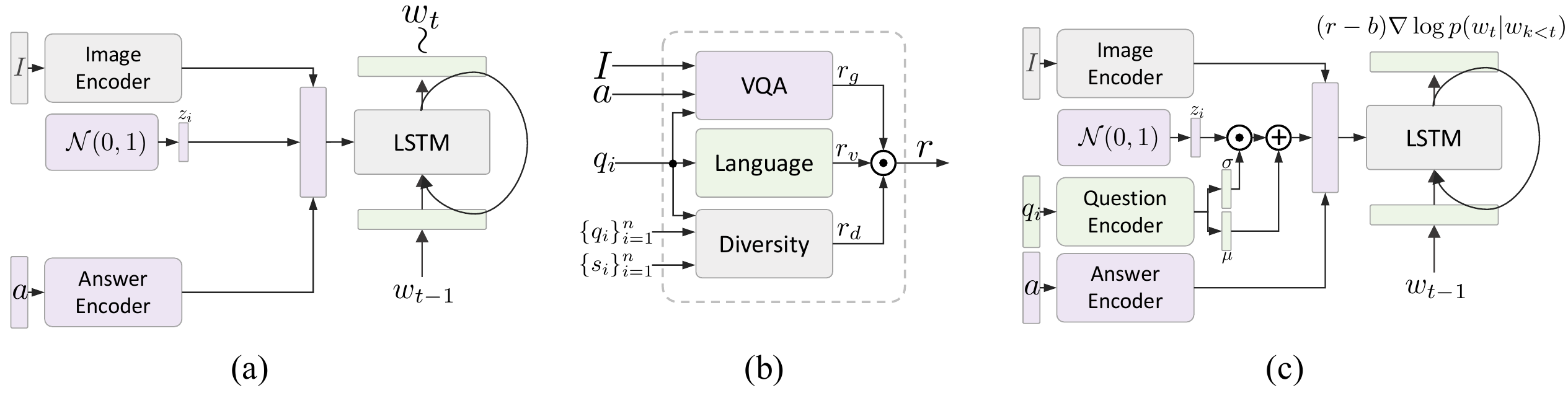}
\caption{Schematic of the proposed  belief set generation framework. There are three modules: (a) Sampling questions. Multiple questions are sampled from an iVQA model by varying $\fnote{\bm{z}}$, which is drawn from a normal distribution. (b) Reward computation. The module takes image $I$, sample questions $\{q_i\}_{i=1}^n$, log probabilities for each path $\{s_i\}_{i=1}^n$ as inputs, and returns the reward $r(q_i)$ for each candidate $q_i$ using a given VQA model under diagnosis. (c) Policy gradient based update. The gradient is passed to the encoder and the decoder of the iVQA model to update the parameters of both.}\label{fig:frame}
\vspace{-0.2cm}
\end{figure*}

In this section we apply our iVQA framework to develop a novel method to diagnose existing VQA models by revealing their \emph{belief sets}. The belief set of a VQA model is defined as a set of visual question-answer pairs which the model believes to be true in the context of the image $I$. The conventional way to evaluate VQA models is to compute their prediction accuracy on a held out testing set of questions and answers. However, this approach is closed-world in the sense that we only know how well the model fits the testing set: which might not be representative of the model's sanity in practice if the VQA dataset is biased, as they are in practice \cite{VQA2.0}. Our belief set based approach is more open-world in that it reveals the set of things that the model believes to be true, whether or not they are annotated in the test set (see Fig.~\ref{fig:frame}). Examining the questions in belief sets thus provides a novel way to study the strengths and weaknesses of different existing VQA models.

\subsection{Belief sets of VQA models}

The belief set $\mathcal{B}$ of a VQA model $\phi(\cdot)$ with respect to an image-answer pair $(I,a)$ can be written mathematically as:
\begin{equation} \label{eq:bs_def}
\begin{split}
& \mathcal{B}(I,a;\phi) = \{q|\Psi(q;I) = a\} \\
& s.t.\quad \Psi(q;I) = \arg\max_{\hat{a} \in \mathcal{A}} \phi(\hat{a}|I,q),
\end{split}
\end{equation}
{where $\mathcal{A}$ is the set of possible answers. That is, the set of questions for which $a$ is the most likely answer.} 
Note that from the definition, the belief set varies for different $(I,a)$ samples given a VQA model $\phi$, and also varies for different models $\phi$ given a particular sample $(I,a)$. So the overall belief set of a model about an image $\mathcal{B}(I;\phi)$ aggregates over answers as:
$\mathcal{B}(I;\phi)=\{(q,a)|\arg\max_{\hat{a}}\phi(\hat{a}|I,q)=a\}$.

Suppose that we have an oracle VQA model $\phi_{ora}(q,I)$, which will assign a confidence score of $1$ for ground truth answer $a$, and $0$ for any other answer given any image $I$ and question $q$. Then a VQA \emph{dataset} can be seen as a sampled subset of the complete belief set of the oracle VQA model (since VQA datasets are not densely annotated). Moreover, the  procedure described in Sec. \ref{sec:ivqa} can be seen as training an approximate belief set generation model for the  VQA oracle via supervised learning.

Having a supervised-learned iVQA model is a good starting point for belief set construction -- the oracle (human annotator) provides data for training a VQA model by providing a sample of its belief set to train the VQA model. But the oracle may not sample its beliefs uniformly, producing a biased dataset. The behaviour of a real VQA model will be different from the oracle: both in making mistakes and  learn from these systematic biases. To visualise and compare existing VQA models we are interested in developing methods to extract the belief set encoded in a given VQA model's parameters. 

\subsection{Training iVQA for VQA belief set extraction}
\label{sec:training iVQA for belief}
To investigate the beliefs of a specific VQA model {$\phi$}, we propose a reinforcement learning (RL) based strategy to train a question generation policy $\pi$. The motivation is similar to that behind adopting an RL strategy for robot mapping and navigation: a given VQA model can be considered as an unknown environment and the iVQA model acts like an autonomous robot which is trained to probe and map the environment (extracting its belief set). {We start with the pretrained policy $\pi_{ora}$, which is obtained by training a variational iVQA model via supervised learning as in Section~\ref{sec:iVQAMethod}. Conceptually, as mentioned above, this is because the supervised iVQA model captures the belief set of the oracle VQA model which is in turn used  to train the VQA mode under diagnosis.  Furthermore, it is necessary computationally: The searching space of questions is extremely large making it difficult to train policy $\pi$ from scratch, while policy $\pi_{ora}$ provides a reasonable starting point.}  

More specifically, the objective is to train an iVQA model $\pi^\phi$ with the objective of generating visual questions that describe the beliefs of $\phi$ (Eq.~(\ref{eq:bs_def})). 
{The training process is shown in Fig.~\ref{fig:frame} and can be described as follows}: Given a randomly sampled vector $\fnote{\bm{z}}$ and image-answer pair $(I, a)$, a question $q^{\prime}$ is sampled based on the current {iVQA} policy $\pi_{\theta_1}$ (Fig.~\ref{fig:frame}(a)). The sampled question $q^{\prime}$ is then evaluated using the VQA model $\phi$ that is being diagnosed (Fig.~\ref{fig:frame}(b)). The returned reward $r$ is then used to update the  parameters of the iVQA model $\pi^\phi$ using policy gradient:
\begin{equation} \label{eq:bs_rl}
{(r-b)\nabla\log p_{\theta_1}(w_t|w_{k<t};\fnote{\bm{z}},I,a)},
\end{equation} 
where $b$ is the baseline to reduce variance (Fig.~\ref{fig:frame}(c)). Note that we also update the encoder part $q_{\theta_2}(\fnote{\bm{z}}|I,q)$ using the back-propagated gradient, which we found is crucial to maintain question generation diversity.

\vspace{0.1cm}
\keypoint{Rewards}
The evolution of the policy is determined by the reward. We consider three factors in designing the reward function: goal orientation, diversity, and legality, {as shown in Fig.~\ref{fig:frame}(b). They are combined in a multiplicative way, because goal orientation is meaningful only when diversity and legality preconditions are met. That is: }
\begin{equation} \label{eq:reward}
r = r_{g} \cdot r_{d} \cdot r_{v},
\end{equation}
{where $r_{g}$, $r_{d}$, $r_{v}$ are goal orientation, diversity, and legality reward components respectively.}

The \textit{goal orientation reward} is to maximise the VQA score of $a$ under {$\phi$} given the sampled question $q'$. It is defined as the score:
\begin{equation}
r_g(q_i) = \phi(a|I,q').
\end{equation}
An iVQA policy that maximises this reward generates questions that follow the beliefs of the VQA model $\phi$. \Ie, Questions for which $\phi$ will predict the given answer with high confidence. However, if $r_g$ is used alone, the RL training would generate meaningless sentences as there is no constraint on the diversity or legality of visual questions.

The \textit{diversity reward} is related to the efficiency of the belief set construction. It encourages different noise vectors in our variational iVQA model to generate distinct questions, by penalising duplicate questions. Assume questions $\{q_i\}_{i=1}^{n}$ are sampled for an image-answer pair $(I,a)$ under $n$ different noise vectors $\{\bm{z}_i\}_{i=1}^{n}$, and the likelihood associated with each candidate is denoted as $\{s_i\}_{i=1}^{n}$. The diversity reward for candidate $q_i$ is computed as:
\begin{equation}
\begin{split}
r_d(q_i) & = \mathbbm{1}(s_i > s^{*}) \\
s.t.~s^{*} & = \max(\{s_i|q_j=q_i,j \neq i \}),
\end{split}
\end{equation}
{which is a \emph{winner-takes-all} strategy: Only the candidate with the maximum likelihood will be given reward. It thus forces the generator to generate diverse questions to seek a large reward.}

The \textit{legality reward} ensures semantic rationality, sentence completeness, and grammatical correctness of  sampled questions. This is non-trivial because off-the-shelf language checkers, \eg, language tool\footnote{\url{https://languagetool.org/}}, can only identify very simple grammar mistakes, so do not serve our purpose. To overcome this problem, we train a \fnote{CNN based language model $p(q)$ \cite{seq_gan}} to perform this task. {It is trained by treating real visual questions as positive samples and randomly perturbed ones as negative samples. We also use part of the sampled questions as negative samples, but a filtration step is applied to remove the random questions which are also in the set of real questions to avoid introducing false negatives. }
The final legality reward is defined as whether the language score is larger than a certain threshold $\epsilon$:
\begin{equation} \label{eq:reward_lm}
r_v(q_i) = \mathbbm{1}(p(q_i)>\epsilon).
\end{equation}
With the proposed training algorithm and reward scheme, we can train a diverse visual question generator that  samples the belief set of a provided VQA model. 

\subsection{Belief set construction}
Once trained, the iVQA model  can sample questions from the belief set of a given VQA model $\phi$ in a trial-and-error manner. Specifically, the iVQA encoder is removed (Fig.~\ref{fig:frame}(b)). Then, given an  image-answer pair $(I,a)$, we sample noise vectors $\bm{z}\sim\mathcal{N}(\bm{0},\bm{1})$. For each noise vector $\fnote{\bm{z}}$ a question $q^{\prime}$ can be sampled. Optionally, its legality can be checked via a language model $\kappa$. Legal questions will then be sent to the VQA model $\phi$ for verification that they are indeed in the belief set: that is, given question $q^{\prime}$, $\phi(a|I,q^{\prime})$ is the highest among all answers in the context of $I$.  The sampling process will be repeated $n$ times to collect enough samples. The above process is summarised in Algorithm~\ref{alg:hack}, {and this whole algorithm can be repeated for different input answers sampled from an answer pool.}

\begin{algorithm}[t]
\SetKwInOut{Input}{input}
\SetKwInOut{Output}{Output}
\SetKwInOut{Init}{Init}

\textbf{Input: } image $I$, answer $a$, VQA model $\phi$, number of trials $n$, and language model $\kappa$ (optional)\;
\textbf{Output: } Belief set $\mathcal{B}(I,a;\phi)$\;
\textbf{Initialise: }{$\mathcal{S}(I,a;\phi)=\emptyset$}\;
\For {$iter=1:n$} 
{
sample vector $\bm{z}_i \sim \mathcal{N}(\bm{0}, \bm{1})$\;
sample question $q'\sim p_{\theta}(q|\fnote{\bm{z}}_i,I,\hat{a})$ using Eqs. (\ref{eq:fuse},~\ref{eq:dec})\;
\If{$\kappa(q^{\prime})=0$} {continue}
\If{$\arg\max_{\hat{a}}\phi(\hat{a}|I,q')=a$}
{$\mathcal{B}(I,a;\phi) \leftarrow \mathcal{B}(I,a;\phi) \bigcup \{q'\}$\;}
}
\caption{Sampling VQA belief sets} \label{alg:hack}
\end{algorithm}

\subsection{VQA model evaluation and diagnosis}
\begin{figure*}[!ht]
\scriptsize
\begin{center}
\begin{tabular}{@{}l@{}|l@{}|l@{}|l}
\toprule
\multicolumn{1}{c}{\textbf{Rephrasing}}	& \multicolumn{1}{|c}{\textbf{Complementary}}	& \multicolumn{1}{|c}{\textbf{Adversarial}} 	& \multicolumn{1}{|c}{\textbf{Irrelevant}} \\ 
\bottomrule
\includegraphics[width=0.22\linewidth]{figs/bs_eg_rep.png} \
&\ \includegraphics[width=0.22\linewidth]{figs/bs_eg_cmp.png} \
&\ \includegraphics[width=0.22\linewidth]{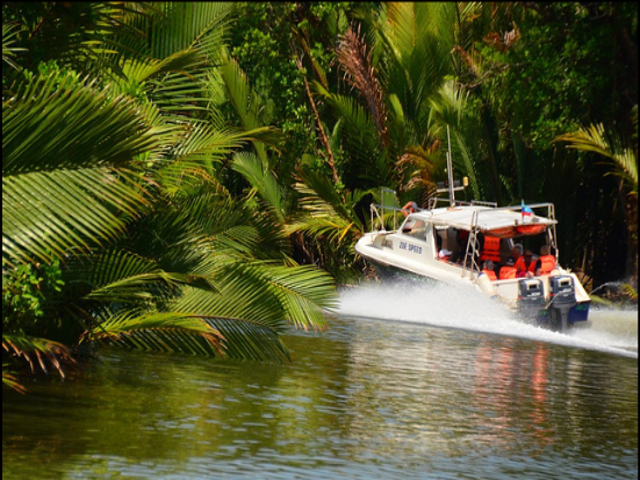} \
&\ \includegraphics[width=0.22\linewidth]{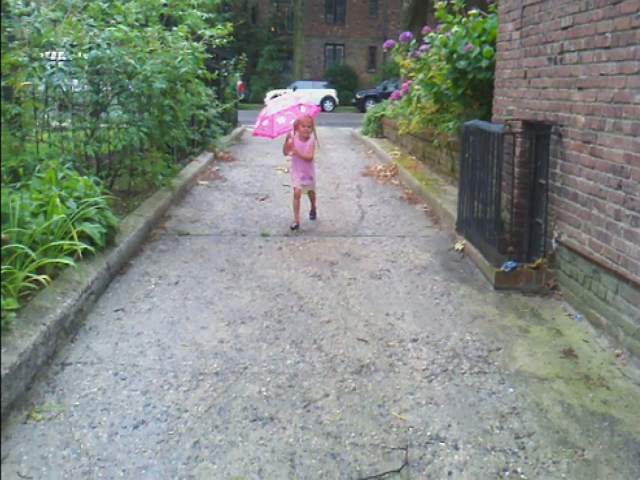} \\
\midrule
GT:\quad What kind of animals are these?  
&GT:\quad What color is the bat?
&GT:\quad How many motors does the boat have?
&GT:\quad What is the girl holding? \\
BS:\quad What animals are these? 
&BS:\quad What color is the batter's helmet?
&BS:\quad How many boats are these?
&BS:\quad What is on the bike? \\
AN:\quad Sheep. &AN:\quad Black. &AN:\quad 2. &AN:\quad Umbrella. \\
\bottomrule
\end{tabular}
\caption{Examples of belief set categorisation. Note that GT stands for ground truth question; BS denotes the question in the belif set; and AN is the answer given the image and GT.}. \label{fig:bs_example}
\end{center}
\end{figure*}
Now given an existing trained VQA model and its belief set extracted using our iVQA model, we explain how to utilise the belief set for evaluating and diagnosing the VQA model.

\keypoint{Belief set annotation} The first step is to annotate the belief set for interpretation.   We aim to categorise the questions into different types. Concretely there are four basic belief types: rephrasing, complementary, adversarial, and irrelevant. \textbf{Rephrasing:} Questions that correspond to visual facts already present in the dataset, but phrased in a different manner. \textbf{Complementary:} {Questions that reflect the image content and also match the given answer, but are not included in the human annotated questions in the VQA training dataset.} \textbf{Adversarial:} These also do not correspond to visual facts in VQA training questions as in the complementary type, but the given answer is incorrect for these questions. These are mistakes, failures of VQA generalisation. \textbf{Irrelevant:} These questions are either false-premise or completely irrelevant to the image or the answer. Examples of the four belief types are shown in Fig.~\ref{fig:bs_example}. A good VQA model should have a large portion of rephrasing and complementary beliefs: since rephrasing beliefs reflect  robustness to linguistic variability in question phrasing, and complementary beliefs show a model's generalisation ability and indicate that it indeed understands the image. In contrast, adversarial and irrelevant beliefs reveal the weaknesses of the VQA model, as both are results of incorrect understanding of  the visual and/or language inputs. 

\keypoint{VQA model evaluation} \fnote{ VQA models can be compared by annotating the belief types in their belief set. 500 belief set examples are manually examined for each model (please see the Supplementary Material for annotation procedure and details).}  The proportions of correct (complementary, rephrasing) versus wrong beliefs (adversarial, irrelevant) provide a  single criterion to compare VQA models.

\keypoint{VQA model diagnosis} To get a better picture of the successes and failures of generalisation of each VQA model, we can further compare the proportion of each belief type. A larger portion of rephrasing indicates better  linguistic understanding and robustness to language perturbation; and larger proportion of complementary questions means generalisation to new visual facts not annotated in the dataset. More importantly, We can also perform qualitative diagnosis by drawing and visualising questions from the belief set to understand the successes and failures of generalisation of each VQA model. {Through the incorrect questions, we can understand  what mistakes/wrong beliefs the model has and further gain insights into why such mistakes are made, e.g., is it particularly poor at detecting certain visual concepts or understanding certain questions? This could then lead to development of potential remedies.}

\section{Experiments on iVQA Diagnosis}

\subsection{Settings}
\keypoint{Implementation Details} Recall that we have two ways to generate potential beliefs: the unconditional (supervised) iVQA approach from Section~3, and the model-conditional (RL) belief set generation approach from Section~5. To more efficiently generate large enough belief sets, we combine the candidate beliefs from both and use the algorithm described in Algorithm \ref{alg:hack} to generate the true belief sets for each model. We use 100 random noise vectors for each image-answer pair for belief set generation in quantitative analysis to ensure feasibility, and sample 1000 random vectors in qualitative results to visualise diverse questions. We conduct ablation study to show that the proposed RL framework (Sec.~\ref{sec:training iVQA for belief}) using all three rewards enables us to efficiently extract the belief set of any given trained VQA model (see the Supplementary Material for details).

\keypoint{VQA Models Selected for Diagnosis} We apply our belief set based diagnosis approach to evaluate and diagnose a number of representative existing VQA models. Specifically, three different VQA models trained with two different VQA datasets (VQA1.0 \cite{vqa_dataset} and VQA2.0 \cite{VQA2.0}) are selected:

\vspace{0.1cm}
 
\noindent \textbf{Vanilla:} This is a standard VQA model without visual attention. Its architecture is similar to  DeeperLSTM+Norm I  introduced in \cite{vqa_dataset}, but instead of  deep LSTM as the sentence encoder, the pretrained skip thought vector \cite{kiros2015skipThought} is used. The image feature is the average pooled \verb|res5c|  of ResNet 152.

\vspace{0.1cm}
 
\noindent \textbf{MLB-attention:} The MLB-attention model \cite{mlb} is a {state of the art} VQA model where soft attention mechanism is used. The skip thought vector is used as the question encoder, and the 14$\times$14 \verb|res5c| feature of ResNet 152 provides the image representation. 

\vspace{0.1cm}

\noindent \textbf{MLB2-attention:} The MLB-attention model retrained on the VQA 2.0 dataset \cite{VQA2.0}.

\vspace{0.1cm}
\noindent \textbf{N2NMN:} The N2MN model \cite{n2nmn} is {another state of the art VQA model with} explicit network structure reasoning. A dynamic inference graph is generated for every question by the layout predictor, while the executer predicts the answer based on the inference graph. The 14$\times$14 \verb|res5c| ResNet 152 provides the image feature, while an LSTM network encodes the sentence. As the code and data are released\footnote{\url{https://github.com/ronghanghu/n2nmn}}, we directly take them for evaluation.

Except MLB2-attention, all the models above are trained on the trainval split of the VQA 1.0 dataset \cite{vqa_dataset}.

\subsection{Results on VQA model evaluation and diagnosis} \label{sec:vis_multi}

\subsubsection{Quantitative results}

\begin{table}[t]
\footnotesize
\begin{center}
\begin{tabular}{@{} l|ccc|c @{}}
\toprule
 	& Vanilla	&  N2NMN	&  MLB-att	&  MLB2-att \\ 
\midrule 
 Rephrase ($\uparrow$)	& 7.8	& 7.6	& { \bf 11.6}	& 11.2 \\ 
 Complementary ($\uparrow$)	& 40.8	& 47.6	& 47.2	& { \bf 54.2} \\ 
 Adversarial ($\downarrow$) & 33.0	& 27.4	& 24.0	& { \bf 19.8} \\ 
 Irrelevant  ($\downarrow$) & 18.4	& 17.4	& 17.2	& { \bf 14.8} \\ 
\midrule
 Correct ($\uparrow$) & 48.6 & 55.2 & 58.8 & { \bf 65.4}\\
 Incorrect ($\downarrow$) & 51.4 & 44.8 & 41.2 & { \bf 34.6}\\
\midrule
 VQA accuracy ($\uparrow$) & 55.30 & { \bf 64.90} & 64.59 & 62.19 \\
\bottomrule
\end{tabular}
\caption{Belief set compositions of different VQA models. Note that the VQA accuracy is evaluated on the test-dev split of VQA 1.0 in all experiments.} \label{tab:compo}
\end{center}
\vspace{-0.3cm}
\end{table}


The belief set  compositions of different VQA models are summarised in Table \ref{tab:compo}, from which we can make the following observations: 
(1) Complementary and adversarial beliefs are the most common ones in all VQA models. Irrelevant beliefs are ranked the 3rd, { indicating the importance of introducing a rejection mechanism into VQA models.} Rephrased questions are the least common, {indicating that our question generator extracts novel beliefs and does not just rely on perturbing known facts.}
(2) MLB-att with visual attention generalises better than the vanilla VQA model. It has more correct  (complementary, rephrase) and fewer incorrect (adversarial, irrelevant) beliefs.
(3) The N2NMN model with explicit reasoning performs similarly to the  attention based MLB-att, and their percentages of correct beliefs are almost the same. However, there are some slight differences: The N2NMN model is better at generalising to complementary visual facts but less robust to adversarial beliefs. 
(4) We can fix a model (MLB-att) and compare the benefit of training on the more balanced VQA 2.0 dataset (MLB2-att vs. MLB-att). The percentage of correct beliefs increased by \fnote{6.6\%}, due to an increased proportion of complementary beliefs. This indicates that the VQA2.0 trained model recognises some new visual facts which it originally missed when trained on VQA 1.0; and to a decreased proportion of adversarial beliefs -- indicating that the VQA2.0-trained model is harder to fool than the original model. This shows that the richer annotation in VQA2.0 is beneficial, however the large outstanding proportion {(\fnote{19.8\%})} of adversarial examples in the belief set indicate that the VQA2.0 dataset is still far from being adequate to train a model that does not suffer from many misconceptions. (5) \doublecheck{Finally, we observe a difference in the standard metric (VQA accuracy) vs our metric (correct beliefs): The standard metric ranks N2NMN and MLB-att slightly above MLB2-att, while our metric ranks MLB2-att clearly above the others. We interpret this as the VQA2.0-trained model actually being better, but the VQA1.0 trained models appear to be better under the standard metric due to overfitting to the biases of this dataset}.

\begin{sidewaystable*}[htbp]
\begin{center}
\scriptsize
\begin{tabular}{@{} l|l|l|l|l @{}}
\toprule
& \multicolumn{1}{c}{Vanilla}	& \multicolumn{1}{|c}{MLB-attention}	& \multicolumn{1}{|c|}{MLB2-attention} 	& \multicolumn{1}{c}{N2NMN} \\ 
\bottomrule
\midrule 
\multirow{12}{*}{\includegraphics[width=0.14\linewidth]{figs/COCO_val2014_000000096549.jpg}}
& is this plane equipped with cruise missiles? (1.00)	& are there vapor trails? (1.00)	& does the airplane have lights on? (0.99)	& is there a lighthouse? (9.51) \\ 
& are these fighter planes? (1.00)	& is there a watermark on the photo? (1.00)	& are the traffic lights on? (0.98)	& does it look cold outside? (9.31) \\ 
& are these commercial airlines? (1.00)	& is it evening? (1.00)	& is it night time? (0.98)	& do you see a board? (9.30) \\ 
& is this a maintenance worker? (1.00)	& can you see any power lines? (1.00)	& is this in an airport? (0.95)	& could this be an intersection? (9.08) \\ 
& are these tourists? (1.00)	& is this plane ascending? (1.00)	& could this be an intersection? (0.95)	& do the mountains have snow on them? (9.08) \\ 
& is it about to storm? (1.00)	& do you see any power lines? (1.00)	& are the planes in an airport? (0.93)	& is the an airport? (9.01) \\ 
& are there vapor trails? (1.00)	& could this be an intersection? (1.00)	& does the plane have stripes? (0.93)	& does the airplane have lights on? (8.90) \\ 
& is this a Japanese airplane? (1.00)	& is it night time? (1.00)	& is this a professional jet? (0.93)	& is there a watermark on the photo? (8.79) \\ 
& is there a watermark on the photo? (1.00)	& can you see mountains? (0.99)	& is there a jet in the picture? (0.92)	& is there any snow? (8.77) \\ 
& was this picture taken from the ground? (1.00)	& does the plane have shadows? (0.99)	& is the an airport? (0.92)	& is this a parking lot? (8.77) \\ 
& is this a museum? (1.00)	& is it getting dark? (0.99)	& is this a professional plane? (0.92)	& is there a forest nearby? (8.76) \\ 
& is there a pilot on the plane? (1.00)	& does the airplane have headlights on? (0.99)	& is this a airport? (0.91)	& is this a bar? (8.68) \\ 
& does the airplane have headlights on? (1.00)	& is this an urban setting? (0.99)	& do you see a plane? (0.91)	& is there a fighter jet in the picture? (8.67) \\ 
Q: Is the landing gear down? & could this be an intersection? (1.00)	& is this a fighter jet? (0.99)	& is this an intersection? (0.91)	& are there stop lights? (8.67) \\ 
A: Yes. & can you see mountains? (1.00)	& is this plane equipped with cruise missiles? (0.99)	& are there any trees visible? (0.91)	& has the plane just landed? (8.59) \\ 

\midrule 
\multirow{12}{*}{\includegraphics[width=0.14\linewidth]{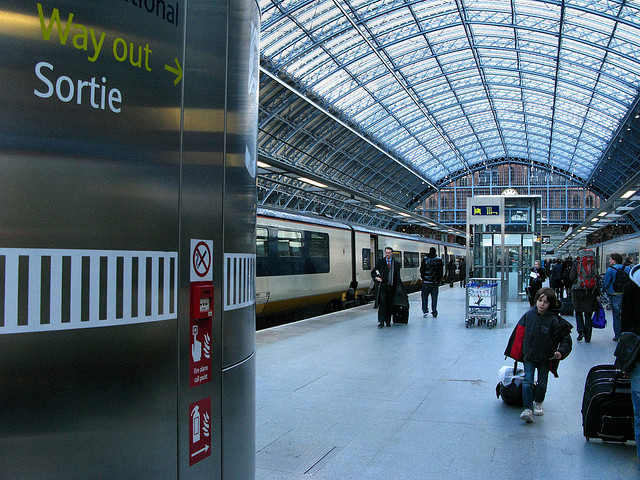}} & was this photo taken before 1980? (1.00)	& is this picture taken in a foreign country? (1.00)	& is there a bridge in the background? (0.98)	& is there a street sign? (10.10) \\ 
& is this picture taken in a foreign country? (1.00)	& are there power lines in the photo? (1.00)	& is this in an airport? (0.97)	& is this a foreign city? (10.05) \\ 
& is this in japan? (1.00)	& is this a tourist attraction? (1.00)	& are there any people? (0.97)	& is this an intersection? (9.88) \\ 
& can you see the photographer? (1.00)	& are there cameras in the picture? (1.00)	& is this a paved road? (0.96)	& is there a jet in the photo? (9.70) \\ 
& is this a tourist attraction? (1.00)	& are there stairs in this scene? (1.00)	& is this a tourist destination? (0.96)	& could this be an airport? (9.64) \\ 
& could this be an airport? (1.00)	& is there an exit? (1.00)	& are there any people in this picture? (0.96)	& could this be a gas station? (9.62) \\ 
& is this a construction site? (1.00)	& can you see a reflection? (1.00)	& is there a flag in this picture? (0.96)	& is this a beautiful place? (9.59) \\ 
& is there a bridge in the background? (1.00)	& is someone wearing a hat? (1.00)	& is there a bike in the picture? (0.96)	& has this photo been altered? (9.58) \\ 
& is this in london? (1.00)	& is there a bridge in this picture? (1.00)	& is there any people in the photo? (0.96)	& is this a conference? (9.55) \\ 
& do the people know each other? (1.00)	& is this public transportation? (1.00)	& is this a tourist attraction? (0.96)	& is this a bookstore? (9.36) \\ 
& is the trolley in the background well manicured? (1.00)	& is this picture taken in a city? (1.00)	& is the photo in color? (0.96)	& is there a train in the picture? (9.30) \\ 
& is this a railway station? (1.00)	& is there any ships in the picture? (1.00)	& does this look like an urban area? (0.95)	& is there a passenger train in this picture? (9.26) \\ 
& is there a flag in this picture? (1.00)	& is the photo blurry? (1.00)	& is it a sunny day? (0.95)	& is this a policeman? (9.22) \\ 
Q: Can you see a fire alarm? & does this look like a convention? (1.00)	& is there a lot of people on the sidewalk? (1.00)	& is it daytime? (0.95)	& is someone walking at the bar? (9.16) \\ 
A: Yes. & are these bikes parked? (1.00)	& is this a tourist destination? (1.00)	& is it cold in this picture? (0.95)	& is there a train in this picture? (9.07) \\ 

\midrule 
\multirow{12}{*}{\includegraphics[width=0.14\linewidth]{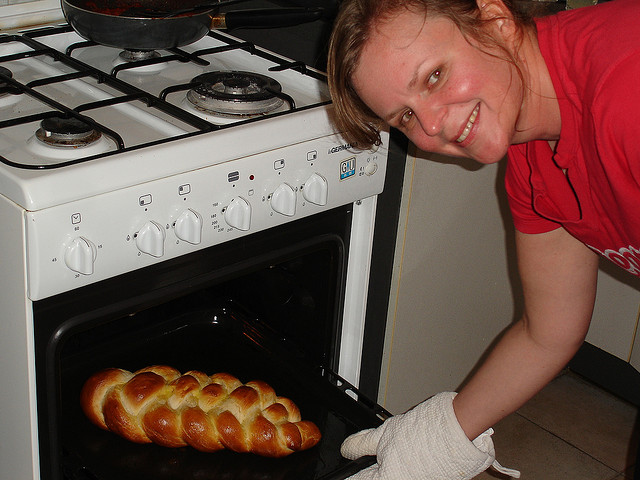}} 
& what is in the bag? (1.00)	& what is in the basket? (0.98)	& what type of food is on the right? (0.96)	& what is holding the food? (8.18) \\ 
& what food is in the basket? (0.99)	& what food is in the basket? (0.97)	& what is in the basket? (0.95)	& what food is in the basket? (8.03) \\ 
& what is in the basket? (0.99)	& what food is on the tray? (0.96)	& what food is in the basket? (0.95)	& what food is the person holding? (7.81) \\ 
& what is in the suitcase? (0.98)	& what type of food is on the table? (0.95)	& what type of food is being prepared? (0.95)	& what food is the woman holding? (7.76) \\ 
& what is on the shelf? (0.96)	& what type of food is that? (0.94)	& what type of food is served on this plate? (0.93)	& what food is in the oven? (7.69) \\ 
& what are they selling? (0.96)	& what type of food is on the right? (0.93)	& what type of food is served? (0.92)	& what food item is on the table? (7.61) \\ 
& what food is on the bag? (0.95)	& what food is being made? (0.92)	& what type of food is on the plate? (0.92)	& what food is on the tray? (7.50) \\ 
& what is on top of the sandwiches? (0.93)	& what food is in the bag? (0.92)	& which type of food is this? (0.92)	& what food is on top of the table? (7.49) \\ 
& what is in the microwave? (0.92)	& what type of food is visible? (0.91)	& what kind of food is on the tray? (0.90)	& what is on the cutting board? (7.39) \\ 
& what food items is on the hot dog? (0.89)	& what type of food is the person holding? (0.89)	& what type of food is in the cart? (0.90)	& what food is next to the sandwich? (7.39) \\ 
& what is on top of the table? (0.85)	& what is being held? (0.89)	& what type of food is it? (0.89)	& what is the food? (7.28) \\ 
& what is in the dish? (0.84)	& what type of food is on the plate? (0.88)	& what type of food is on these plate? (0.88)	& what is laying on the plate? (7.26) \\ 
& what is on the plates? (0.75)	& what type of food is on the front table? (0.87)	& what type of food is that? (0.88)	& what is she cooking? (7.12) \\ 
Q: What kind of food is this? & what is being fried? (0.74)	& what type of food is in the cart? (0.85)	& what kind of food is on the hand? (0.88)	& what does she cutting? (6.97) \\ 
A: Bread. & what is on the napkin? (0.70)	& what kind of food is the woman holding? (0.85)	& what food is on the bag? (0.87)	& what food is in the bowl? (6.89) \\ 

\bottomrule
\midrule 
\multirow{12}{*}{\includegraphics[width=0.14\linewidth]{figs/COCO_val2014_000000096549.jpg}} 
 	& do the planes have propellers? (1.00)	& is the plane upside down? (0.93)	& could this be an intersection? (0.95)	& is this plan visible? (9.16) \\ 
 	& is there a watermark on the photo? (1.00)	& are there mountains in this photo? (0.92)	& are they at an airport? (0.94)	& is there a river in the picture? (8.37) \\ 
 	& is this a commercial airline? (1.00)	& is this a professional plane? (0.92)	& does the plane have a stripe? (0.87)	& are these fighter planes? (8.00) \\ 
 	& are there any buildings in the photo? (1.00)	& is there a river in the picture? (0.91)	& is this a commercial aircraft? (0.87)	& is there a rainbow? (7.97) \\ 
 	& did the plane just land? (1.00)	& can you make a stuff at this location? (0.91)	& does the plane have windows? (0.85)	& does it look like it 's going to rain? (7.96) \\ 
 	& is this a cookie? (1.00)	& is the airplane upside down? (0.90)	& does the plane have a shadow? (0.74)	& is this an urban area? (7.93) \\ 
 	& is this a outdoor plane? (1.00)	& is the plane going to land? (0.86)	& is this a display? (0.69)	& are those fighter planes? (7.53) \\ 
 	& is the pilot on the plane? (1.00)	& is there any snow? (0.81)	& is this a fighter plane? (0.61)	& was this picture taken from the ground? (7.40) \\ 
 	& are these people going towards the same plane? (0.99)	& can you see any milk? (0.76)	& is a landing gear down? (0.58)	& is this a city scene? (7.39) \\ 
 	& are the trees parked? (0.90)	& does the plane have four engines? (0.74)	& is the plane ready to take off? (0.58)	& is this the runway? (7.28) \\ 
 	& does the plane have bat? (0.89)	& is this a family? (0.62)	& are there clouds? (0.58)	& is the plane large? (7.04) \\ 
 	& will the plane plane that when they are out? (0.83)	& is there a pilot of the plane? (0.61)	& does it look like it 's going to rain? (0.55)	& is this a military jet? (6.90) \\ 
 	& does it look crowded? (0.78)	& is the plane near the mountains? (0.59)	& is there a lot of traffic? (0.53)	& is there a man in the picture? (6.81) \\ 
Q: Is the landing gear down? 
 	& is this day time? (0.67)	& does the grass need to be mowed? (0.55)	& is this planes getting ready to take off? (0.52)	& is the sky black? (6.70) \\ 
A: Yes.
 	& is it storming? (0.61)	& are these all women? (0.53)	& are these people facing the same direction? (0.52)	& is this picture in focus? (5.93) \\ 
\bottomrule
\end{tabular}
\caption{Visualisation of the top beliefs of different VQA models. For the top three rows, 15 candidates with maximum VQA scores are selected, while random candidates are selected in the last row.}
\label{tab:bs_example}
\end{center}
\end{sidewaystable*}

\begin{table}[t]
\footnotesize
\begin{center}
\begin{tabular}{@{} l|cccc|c @{}}
\toprule
Models & Vanilla	& N2NMN2	& MLB-att	& MLB2-att	& MCB-att$^{\dagger}$ \\ 
\midrule 
accuracy	& 14.54	& 24.82	& 24.47	& 45.39	& 26.24 \\ 
\bottomrule
\end{tabular}
\end{center}
\caption{VQA accuracy on the proposed extra-challenging VQA dataset, where $\dagger$ indicates the model is not contribute to the set of hard examples.}
\label{tab:vqa_hard}
\vspace{-0.2cm}
\end{table}

\subsubsection{Qualitative results and diagnosis}
The belief sets of the VQA models on three selected image-answer pairs are shown in Table \ref{tab:bs_example}. As there are too many unique visual questions in the belief set to show exhaustively, we select the {beliefs to show} via two protocols: maximum belief (Fig.~\ref{tab:bs_example} Row 1-3) and random (Fig. \ref{tab:bs_example} Row 4). For  maximum belief, we select the beliefs with maximum VQA scores and manually remove questions with the same meaning. For the random protocol, we randomly select from the belief sets questions of different confidence intervals.

The result shows that the VQA models encode many visual facts about an image other than human annotation. For example, besides `landing gear down', they additionally recognise `headlights on', 'night time', `airports' \etc for the aeroplane image (Row 1). Even though there are mistakes, attention models understand the image better as  there are more correct questions in their belief sets. Furthermore, the MLB-attention model performs significantly better than others when trained with VQA 2.0 (MLB2-att), indicating that the rebalancing in VQA 2.0 data helps.

Nevertheless various weaknesses of the VQA models are also revealed through the diagnosis.
Firstly,  semantic structure and common knowledge are poorly represented in the VQA models. For example, an aeroplane cannot be a `fighter plane' and `commercial aircraft' at the same time; places are unlikely to simultaneously be `lighthouse', `airport', `parking lot', `bar' \etc; `ships' usually do not co-occur with `city' or `sidewalk'. However, all of them simultaneously appear in the belief sets of these VQA models given a particular image and an answer.  While this may be due to these kinds of questions being rare in the original training set, these kinds of issues are particularly severe for the N2NMN model  indicating that its generalisation is weak.
Secondly, the correlation between the objects in questions and the answer affects the prediction. For example, as shown in the third row of Table \ref{tab:bs_example}, the false-premise question ``what food is in the basket?'' achieves a score even higher than the true-premise question ``what food is on the tray?''. Strikingly, this happens for every VQA model compared. It indicates that correlation or co-occurrence is over exploited by existing VQA models and thus these models may be making shortcut predictions without fully understanding the image/question. Even re-balancing the data in VQA2.0 is far from fully solving the problem.

\subsection{A challenging VQA dataset} Based on the union of our collected adversarial examples for the four VQA models we evaluated, we release a small but challenging dataset for VQA which is designed to  be a focused extra-challenging benchmark for future VQA studies. \fnote{It contains 282 unique samples over 195 images, and 77, 74, 76, 55 examples are from the adversarial beliefs of Vanilla, N2NMN, MLB-att, and MLB2-att respectively. The correct question-answer pairs are annotated manually for these wrong beliefs for quantitative evaluation. Examples of the datasets can be found in the Supplementary Material.  
The accuracy (correctly predicting the answer given an image and question) of the above VQA models as well as MCB-att\footnote{An attention model with multi-modal compact bilinear pooling as the fusion strategy \cite{mcb}. \url{https://github.com/akirafukui/vqa-mcb}}  on the new dataset are shown in Table \ref{tab:vqa_hard}. The results show that the evaluated VQA models (including MCB-att which was not used for belief set generation) all struggle on these extra-challenging  examples. Comparing Table \ref{tab:vqa_hard} with Table \ref{tab:compo}, the accuracy drops as much as 40\% (Vanilla, from 55.30\% to 14.54\%). This dataset can be used to complement existing VQA dataset to evaluate newly proposed VQA models to see that whether they are still making the same mistakes as previous models.  }

\subsection{Insights and future directions}
From our results, we found that inadequate language modelling, concept relationship reasoning,  and dataset bias are the main problems for all VQA models diagnosed using our iVQA model. For example, all models suffers from irrelevant  or adversarial questions. 
These problems have drawn increasing attention and several approaches are proposed to overcome them. One  is to better balance the data to reduce bias: Goyal \etal~\cite{VQA2.0} balance the answer distribution by providing more visual questions for each image, which makes the questions for each topic more balanced thus helping to  learn a more reliable model. Explicit reasoning strategies have also been investigated in \cite{n2nmn,count1}. This can be seen as introducing more prior knowledge in VQA architecture design, which achieves higher performance with a fixed amount of annotation. Additional annotations or models of related tasks are used in \cite{botup,wang2017vqa,NIPS2017_6658} to improve the visual module, interpreting VQA as a combination of multiple sub-problems, \eg, object detection, classification, OCR \etc.  All these methods are useful in improving VQA model performance. However,  they do not fully address the limitations of VQA models revealed in this study. So we suggest the following directions for future VQA research:

\keypoint{Annotation matters} A densely annotated VQA dataset where most visual facts are annotated is needed. Current VQA datasets are too sparse to provide enough supervision to learn a strong visual module, and too biased in their distribution. This is reflected in the large portion of adversarial beliefs. Given that `a picture speaks a thousand words', dense annotation is likely to be very costly. But the ability to reveal model's belief sets means that active learning \cite{settles2012activeLearning,activeVQA} becomes a promising direction to fill the gaps in existing annotations, allowing better models to be trained.

\keypoint{Consistent Beliefs} \doublecheck{Our analysis showed that VQA models engage in \emph{doublethink}: Holding multiple mutually inconsistent beliefs. It may be useful to study multi-task inference \cite{bilen2016mtlInference} models that can reduce this cognitive dissonance \cite{festinger1962cognitiveDissonance}. }

\keypoint{Joint learning of VQA and QRD} Joint learning of visual question answering (VQA) and question relevance detection (QRD) is an important issue. From an application perspective, a VQA model should be able to reject irrelevant questions rather than giving a confident and meaningless answer. From a training perspective, QRD would enforce a model to reason about the correctness of implied visual facts in the image, which provides an additional task for multi-task training.

\section{Conclusion}
We have introduced the novel task of inverse VQA as an alternative multi-modal visual intelligence challenge to the popular VQA paradigm. As a multi-modal intelligence benchmark, our analyses suggest that iVQA is appealing in terms of being less game-able via exploiting label-bias, more clearly requiring the mutual grounding and understanding of both visual and linguistic modalities, and naturally providing an open-world prediction setting. We show that iVQA also provides a novel approach to evaluating and diagnosing VQA models and datasets by revealing the belief sets held by a given trained VQA model.

\ifCLASSOPTIONcaptionsoff
  \newpage
\fi



\bibliographystyle{IEEEtran}
\bibliography{lstm}

\vspace{-1.2cm}
\begin{IEEEbiography}[{\includegraphics[width=1in,height=1.25in,clip,keepaspectratio]{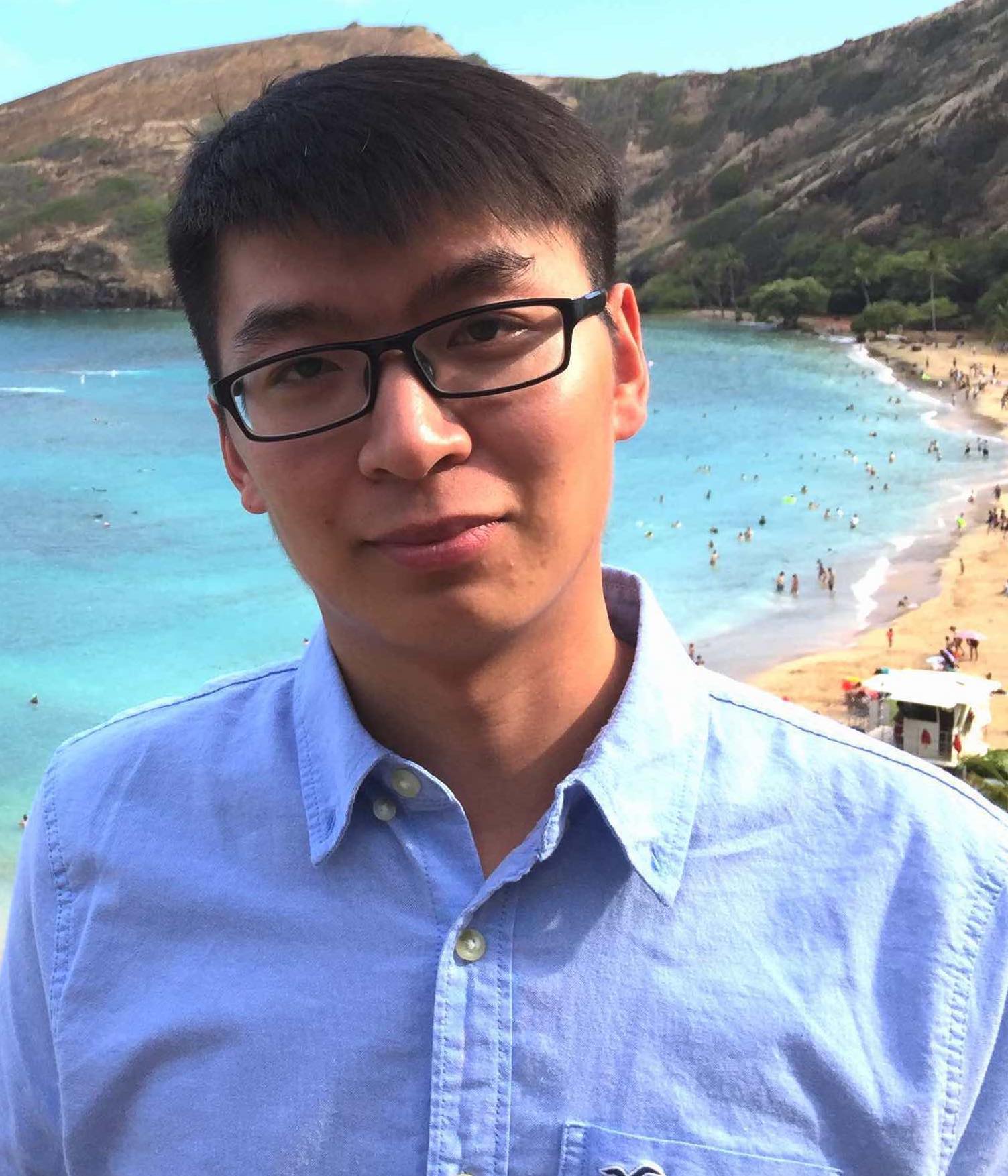}}]{Feng Liu}
is a Ph. D. candidate in School of Automation, Southeast University, China. He received the B.S. degree in Electrical Engineering from  China University of Mining and Technology, Xuzhou, China in 2010. His research interests include computer vision, natural language processing, deep learning, and reinforcement learning.
\end{IEEEbiography}
\vspace{-1cm}

\begin{IEEEbiography}[{\includegraphics[width=1in,height=1.25in,clip,keepaspectratio]{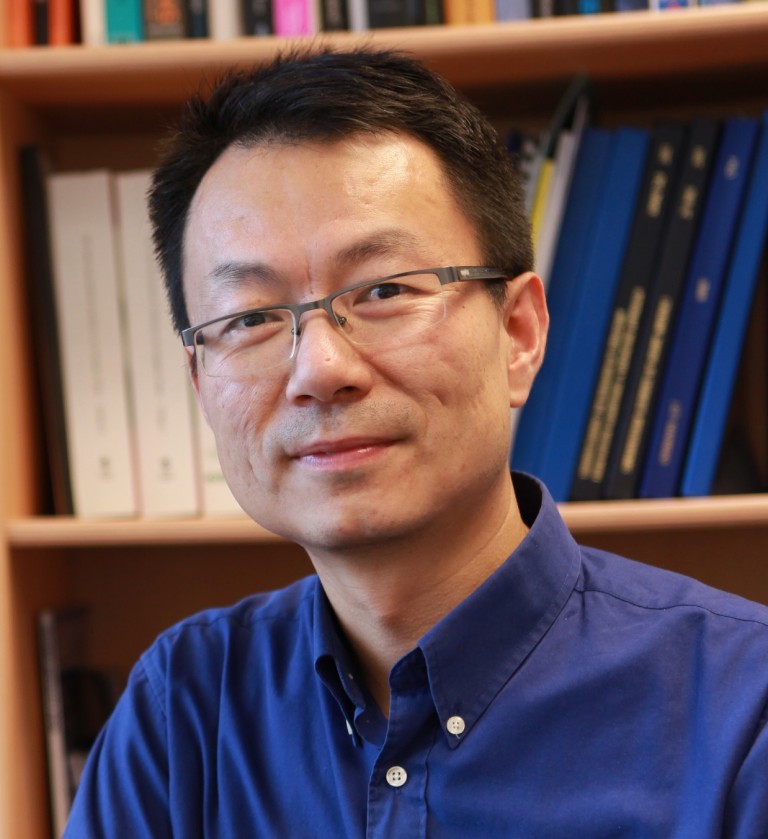}}]{Tao Xiang} received the BS degree from Xi'an Jiaotong University, Xi'an, China, in 1995, and the PhD degree from the National University of Singapore, in 2001. He
is a Reader in Computer Vision and Multimedia, Vision Group, School of Electronic Engineering and Computer Science, Queen Mary University of London. His interests include computer vision, pattern recognition and machine learning. 
\end{IEEEbiography}
\vspace{-1cm}

\begin{IEEEbiography}[{\includegraphics[width=1in,height=1.25in,clip,keepaspectratio]{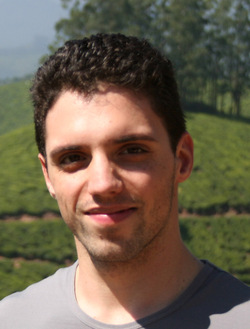}}]{Timothy M. Hospedales}
is a Reader within IPAB in the School of Informatics at the University of Edinburgh, and Visiting Reader at Queen Mary University of London. His research focuses on machine learning, particularly life-long, transfer and active learning, with both probabilistic and deep learning approaches. He has studied a variety application areas including computer vision, language, robot control and beyond.
\end{IEEEbiography}
\vspace{-1cm}

\begin{IEEEbiography}[{\includegraphics[width=1in,height=1.25in,clip,keepaspectratio]{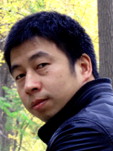}}]{Wankou Yang}
is an associate professor in School of Automation, Southeast University, Nanjing, China. He received the B.S., M.S. and Ph.D. degrees in the School of Computer Science and Technology, Nanjing University of Science and Technology (NUST), in 2002, 2004, and 2009 respectively. His research interests include pattern recognition, computer vision and machine learning.
\end{IEEEbiography}
\vspace{-1cm}

\begin{IEEEbiography}[{\includegraphics[width=1in,height=1.25in,clip,keepaspectratio]{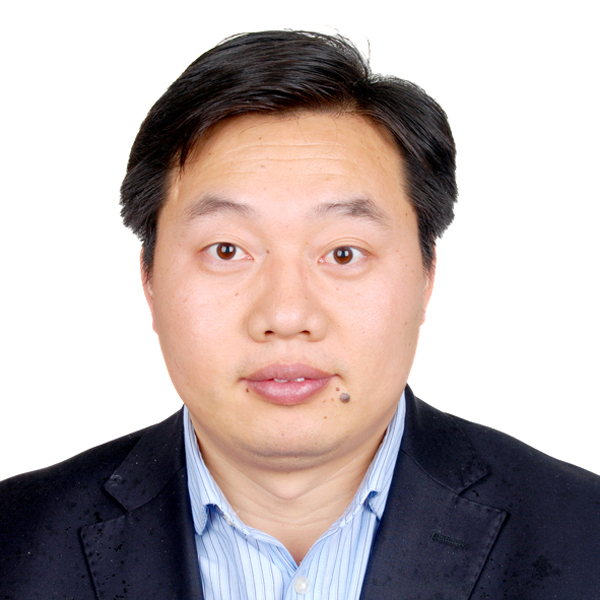}}]{Changyin Sun}
is a professor in School of Automation at the Southeast University, China. He received the M.S. and Ph.D. degrees in Electrical Engineering from the Southeast University in 2001 and 2003 respectively. His research interests include control theory, pattern recognition and machine learning.
\end{IEEEbiography}

\end{document}


\title{Inverse Visual Question Answering: \\A New Benchmark and VQA Diagnosis Tool\\Supplementary Materials}

\author{Feng Liu,
Tao Xiang,
Timothy M. Hospedales, 
Wankou Yang, and
Changyin Sun
}

\maketitle

\IEEEdisplaynontitleabstractindextext

%
\IEEEpeerreviewmaketitle

\section{Annotation procedure of belief types}

As mentioned in Section 5.4 of the main paper,  the belief set of given VQA model to be diagnosed is generated using our RL-trained iVQA model. Subsequently each question-answer-image tuple in the belief set needs to be examined by a human annotator and annotated into four categories for evaluation.   The annotation is collected via the following procedure: For each model, image-candidate-answer tuples are sampled. Then the human labellers are shown an image, its original question-answer pairs in the dataset, and a candidate from the belief set, where the original question is called reference, and the candidate is referred as target. The following questions are asked: \emph{``Is the target relevant to the image?''}, \emph{``Does the target refer to the same visual fact as the reference?''}, and \emph{``Is the answer correct for the target question?''}. Given the above responses, the belief types are assigned following the rules in Table \ref{tab:bs_type}. 500 samples are manually examined to obtain the belief set of each model.

\begin{table}[ht]
\begin{center}
\begin{tabular}{@{} l|c|c|c @{}}
\toprule
Types	& relevance	& visual fact	& correct \\ 
\midrule 
rephrase	& Yes	& Same	& Yes \\ 
complementary	& Yes	& Different	& Yes \\ 
adversarial	& Yes	& Different	& No \\ 
irrelevant	& No	& NA	& NA \\ 
\bottomrule
\end{tabular}
\caption{Mapping from annotator answers to belief types.} \label{tab:bs_type}
\end{center}
\end{table}

\subsection{Validation of the RL training scheme of iVQA for Belief Set Extraction}
{In this section, we perform an ablation study to validate the importance of our proposed reward scheme in the RL framework (Section 5.2 of the main paper) as key for successful training a iVQA model for efficiently extracting the belief set of a given VQA model. We compare different methods for generating candidate beliefs using four evaluation metrics: oracle CIDEr (\textbf{O-C}), BLEU-4 (\textbf{O-B4}), mean VQA score (\textbf{m-S}), and average number of unique questions for each sample (U\textbf{\#}). The oracle linguistic scores are defined as the maximum matching score between a ground truth question and all generated candidate questions, which reflects the amount of image content discovered by the iVQA model. The mean VQA score is the average score returned by the VQA model when tested on the generated candidates. The U\# metric is obtained by averaging the number of candidate questions for every image-answer pair on the testing set. }

{We train the proposed model on the training split with the soft attention model (MLB-att) \cite{mlb} as the VQA model to provide scores for goal oriented reward as well as the m-S metric, and test the above four metrics on the test split of the VQA 1.0 dataset. The results are shown in Table \ref{tab:ablation}. It shows that a model will fail and eventually end up with generating empty questions when legality factor is removed from the overall reward. That is because illegal questions (incomplete sentence, incorrect grammar) can also obtain high VQA scores, so the model will lose its knowledge about generating proper questions and deteriorate gradually in the following training process. After adding the legality check, the incorrect questions could be suppressed and the training can be proceed properly, but many noise vectors tend to generate the same question as diversity is not encouraged in the reward. With the proposed diversity reward, our model  eventually generates less duplicate candidates with a fixed amount of noise vectors, so more unique questions are discovered in the RL training process. Hence it has more opportunity to attack the VQA model. This justifies using our full reward (G+L+D, Eq.~(11) in the main paper) to train our iVQA model for belief set extraction.}

\subsection{Examples of the Proposed Challenging VQA Dataset}

\begin{figure*}[!ht]
\scriptsize
\begin{center}
\begin{tabular}{@{}l@{}l@{}l@{}l}

\includegraphics[width=0.22\linewidth]{figs/hard_eg/COCO_val2014_000000476406.jpg} \
&\ \includegraphics[width=0.22\linewidth]{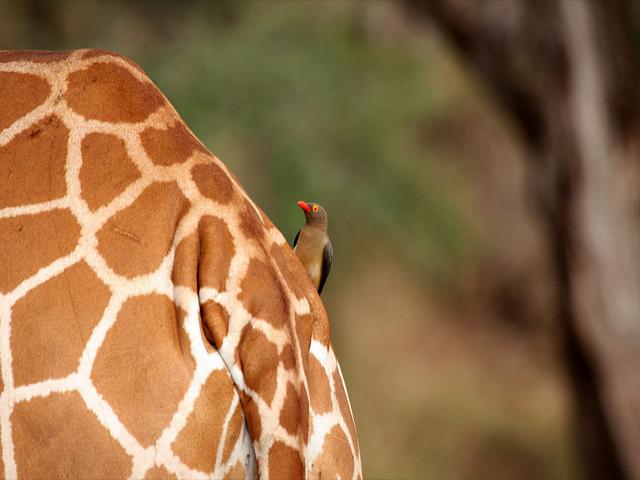} \
&\ \includegraphics[width=0.22\linewidth]{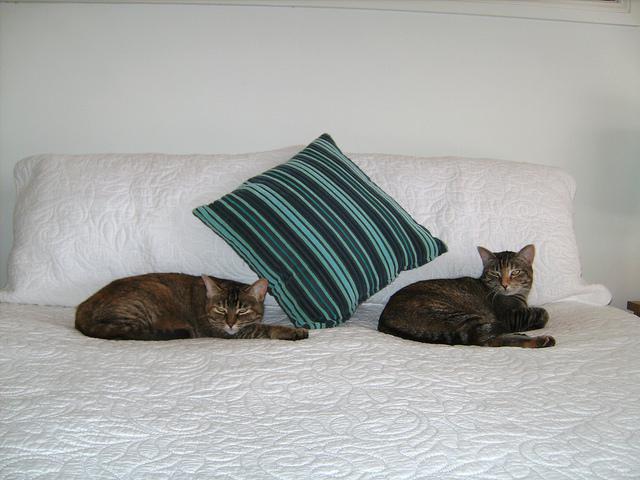} \
&\ \includegraphics[width=0.22\linewidth]{figs/hard_eg/COCO_val2014_000000362275.jpg} \\

Q:\quad How many train tracks do you see?  
&Q:\quad How many giraffes are there?
&Q:\quad How many pillows do you see?
&Q:\quad How many animals are in the picture? \\
A:\quad 4
&A:\quad 1
&A:\quad 3
&A:\quad 3 \\

\includegraphics[width=0.22\linewidth]{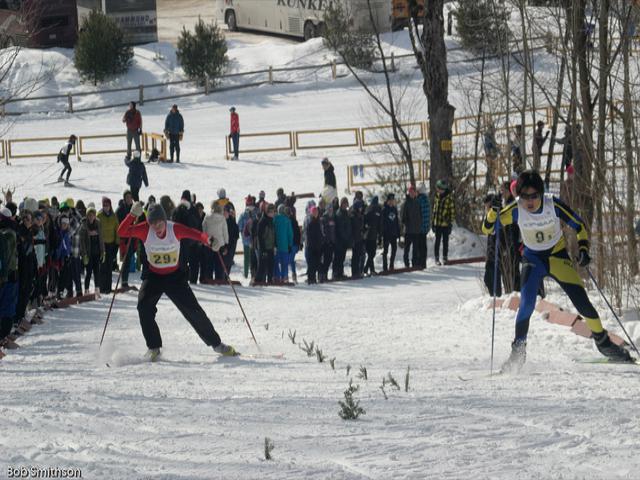} \
&\ \includegraphics[width=0.22\linewidth]{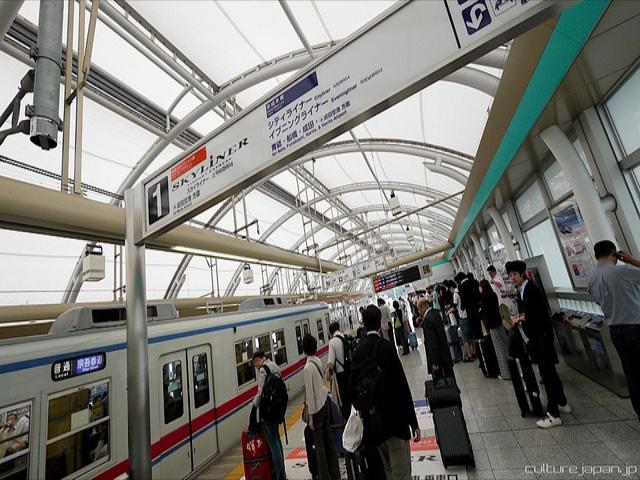} \
&\ \includegraphics[width=0.22\linewidth]{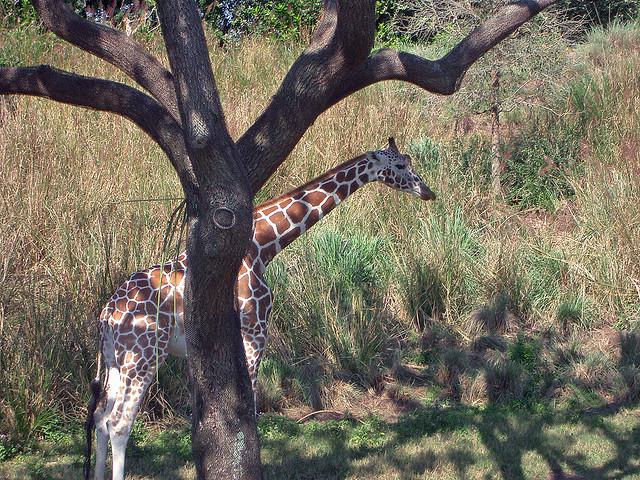} \
&\ \includegraphics[width=0.22\linewidth]{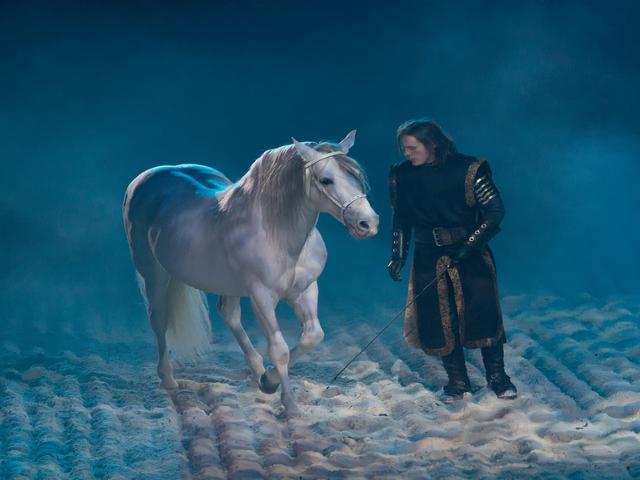} \\

Q:\quad What is the color of the grass?
&Q:\quad Are these people arriving or departing?
&Q:\quad What is behind the giraffe?
&Q:\quad What color are the horses eyes? \\
A:\quad Green
&A:\quad Departing
&A:\quad Grass
&A:\quad Black \\

\includegraphics[width=0.22\linewidth]{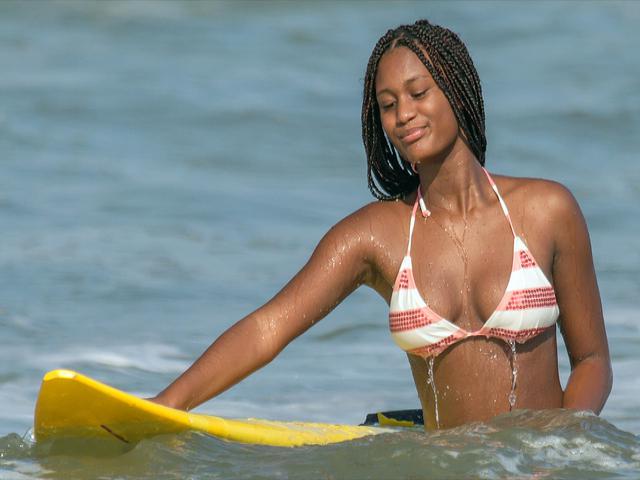} \
&\ \includegraphics[width=0.22\linewidth]{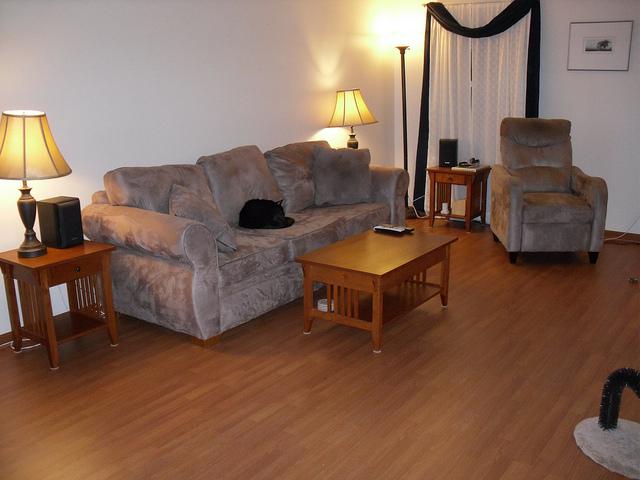} \
&\ \includegraphics[width=0.22\linewidth]{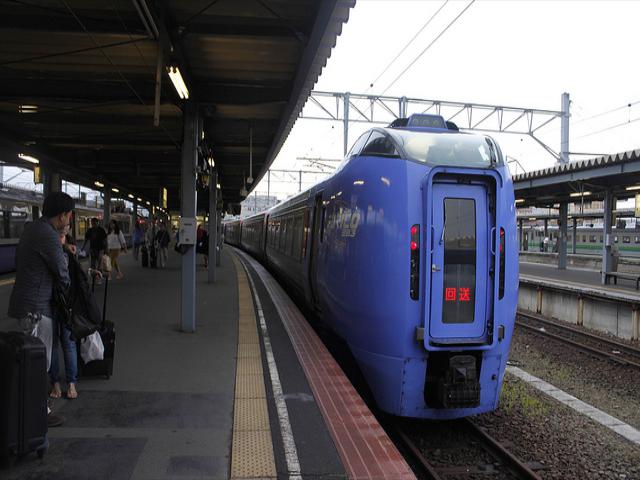} \
&\ \includegraphics[width=0.22\linewidth]{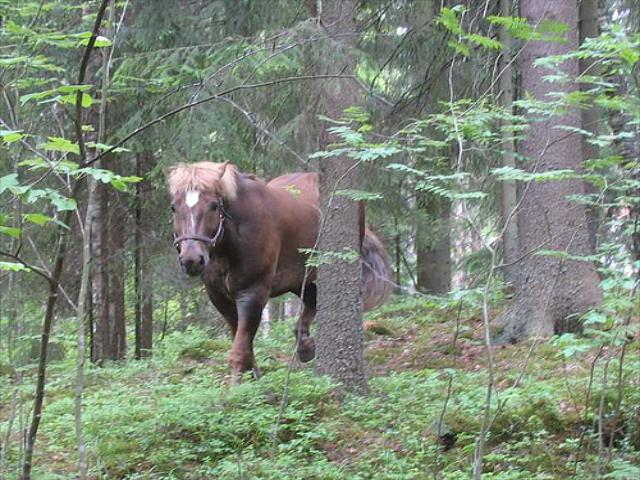} \\

Q:\quad Is she wearing a dress?
&Q:\quad Could this be a hotel room?
&Q:\quad Does the train have a stripe?
&Q:\quad Does the horse have a saddle?   \\
A:\quad No
&A:\quad No
&A:\quad No
&A:\quad No \\

\includegraphics[width=0.22\linewidth]{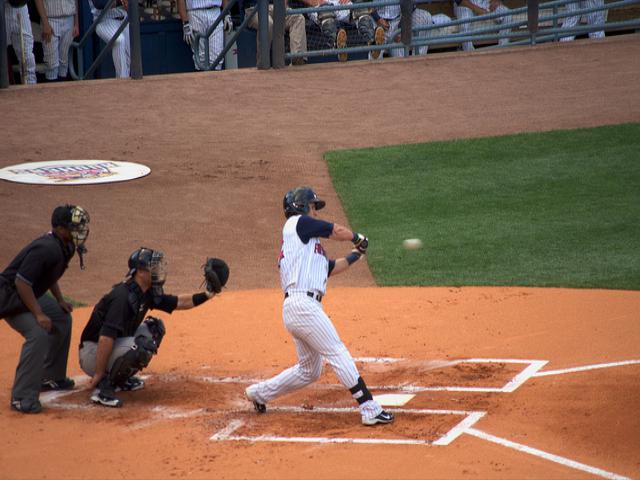} \
&\ \includegraphics[width=0.22\linewidth]{figs/hard_eg/COCO_val2014_000000022892.jpg} \
&\ \includegraphics[width=0.22\linewidth]{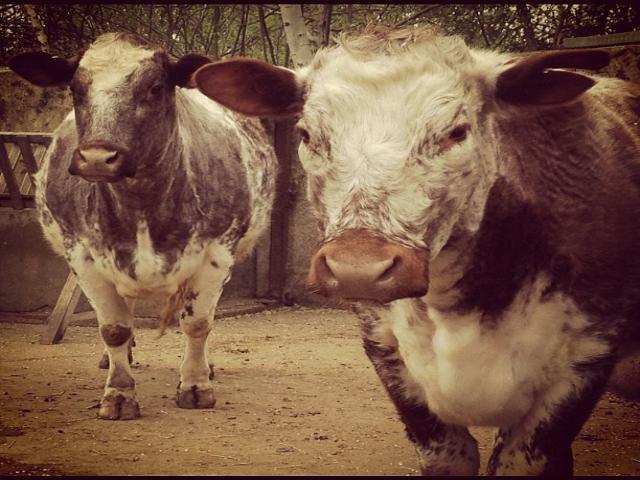} \
&\ \includegraphics[width=0.22\linewidth]{figs/hard_eg/COCO_val2014_000000448871.jpg} \\

Q:\quad Is this picture in color?
&Q:\quad Is there a baby animal in this photo?
&Q:\quad Does the cow have horns?
&Q:\quad Is this an airport?  \\
A:\quad Yes
&A:\quad No
&A:\quad No
&A:\quad No \\
\end{tabular}
\caption{Examples of in the extra-challenging VQA dataset.} \label{fig:hard_example}
\vspace{-0.3cm}
\end{center}
\end{figure*}

In Section 6.3 of the main paper, we introduced an extra-challenging VQA dataset obtained by combining the adversarial beliefs of four VQA models. Some examples of the  datasets are shown in Fig.~\ref{fig:hard_example}.

\begin{table}[t]
\footnotesize
\begin{center}
\begin{tabular}{@{} l|ccc|c @{}}
\toprule
Methods 	& O-C	& O-B4	& m-S	& U\# \\ 
\midrule 
G 	& - & -	&  - & failed \\ 
G+L 	& 1.973	& 0.255	& 0.393	& 10.62 \\ 
G+L+D  & \textbf{2.135}	& \textbf{0.268}	& \textbf{0.451}	& \textbf{14.22} \\ 
\bottomrule
\end{tabular}
\caption{Abliation study of the proposed reward scheme, where G, L, D indicates goal oriented reward, legality reward, and diversity reward respectively.} \label{tab:ablation}
\end{center}
\end{table}

\ifCLASSOPTIONcaptionsoff
  \newpage
\fi



\bibliographystyle{IEEEtran}
\bibliography{lstm}
